\documentclass[runningheads]{llncs}

\usepackage[mobile]{eccv}

\usepackage{eccvabbrv}

\usepackage{graphicx}
\usepackage{booktabs}
\usepackage{multirow}
\newcommand{\Rows}[1]{\multirow{2}{*}{#1}}

\usepackage[accsupp]{axessibility}  

\usepackage[pagebackref,breaklinks,colorlinks,citecolor=eccvblue]{hyperref}

\usepackage{orcidlink}

\begin{document}

\title{Strong and Controllable Blind Image Decomposition} 

\titlerunning{Strong and Controllable Blind Image Decomposition}

\author{ Zeyu Zhang \inst{1}* \and
Junlin Han\inst{2}* \and
Chenhui Gou\inst{3}* \and
\\
Hongdong Li\inst{1} \and
Liang Zheng\inst{1} 
}

\authorrunning{Zeyu Zhang et al.}

\institute{Australian National University \and
University of Oxford \and
Monash University \\
* Equal Contribution\\
\email{zeyu.zhang@anu.edu,au, junlin.han@eng.ox.ac.uk, chenhui.gou@monash.edu, hongdong.li@anu.edu.au, liang.zheng@anu.edu.au  }}

\maketitle
\vspace{-10pt}
\begin{figure}[!htb]
\centering 
\includegraphics[width=1\linewidth]{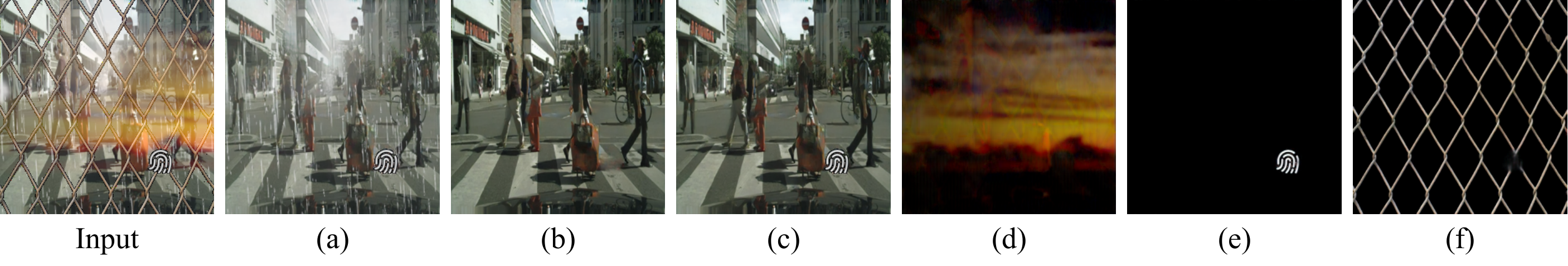}
\caption{\textbf{Varying user demands for image processing.} Whether a component in an image is considered deteriorative or valuable content hinges on the user's intention. For a cluttered input image degraded by various components, users might want to (a) remove the interfering fence and reflection, obtaining a rainy street scene, (b) eliminate all degradations to get a clean street scene, (c) keep the watermark for the sake of copyright protection, and (d$\sim$f) extract the reflection, watermark, and fence.}
\label{fig:control}
\end{figure}
\vspace{-20pt}

\begin{abstract}
Blind image decomposition aims to decompose all components present in an image, typically used to restore a multi-degraded input image.  While fully recovering the clean image is appealing, in some scenarios, users might want to retain certain degradations, such as watermarks, for copyright protection (refer to Figure~\ref{fig:control}). To address this need, we add controllability to the blind image decomposition process, allowing users to enter which types of degradation to remove or retain. We design an architecture named controllable blind image decomposition network. Inserted in the middle of U-Net structure, our method first decomposes the input feature maps and then recombines them according to user instructions. Advantageously, this functionality is implemented at minimal computational cost: decomposition and recombination are all parameter-free.
Experimentally, our system excels in blind image decomposition tasks and can outputs partially or fully restored images that well reflect user intentions. Furthermore, we evaluate and configure different options for the network structure and loss functions. This, combined with the proposed decomposition-and-recombination method, yields an efficient and competitive system for blind image decomposition, compared with current state-of-the-art methods.  Code is available at  \textcolor{red}{\href{https://github.com/Zhangzeyu97/CBD.git}{GitHub}}.
  \keywords{Image Decomposition, Low-level Vision, Rain Removal}
\end{abstract}

\begin{figure*}[htb]
\includegraphics[width=1\linewidth]{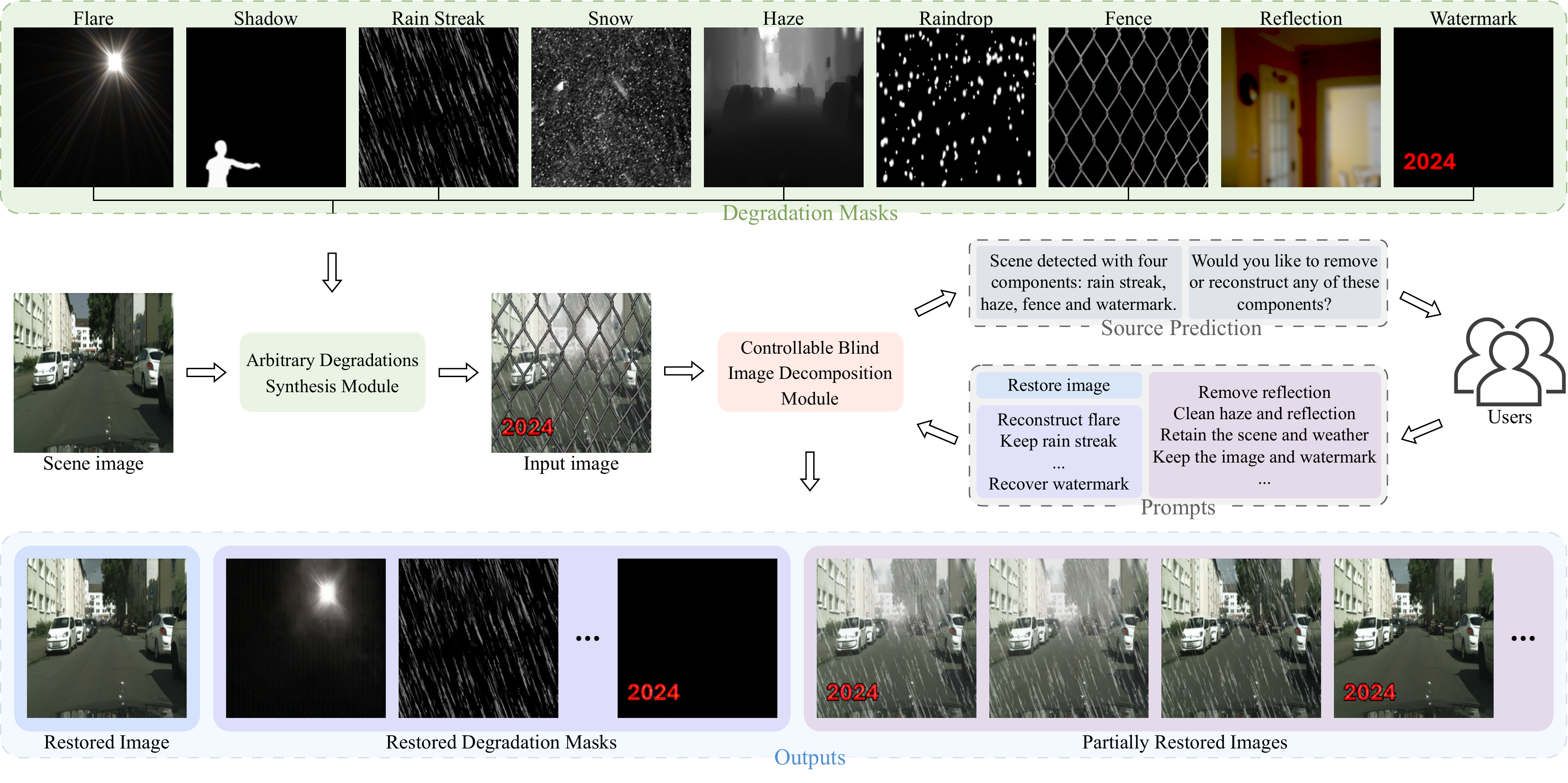}
\caption{\textbf{Workflow of controllable blind image decomposition (controllable BID).} For synthetic input images that exhibit one or multiple degradations, the controllable BID module first predicts what degradations are present. This prediction of components present aids users in making precise instruction prompts.  Once these prompts are formulated, the controllable BID module then processes the input image based on these user-provided instructions. The generated images include the restored image, the degradation masks, and images with specific degradations removed.}
\label{fig:pipeline}
\end{figure*}

\vspace{-20pt}
\section{Introduction}
\label{sec:intro}

Acknowledging the personalized needs for processing images containing multiple types of degradation, it becomes necessary to develop a framework capable of adhering to specific user instructions for targeted image restoration. 
One approach could involve employing multiple specialized models for distinct tasks~\cite{Kupyn_Martyniuk_Wu_Wang_2019, Dong_Loy_He_Tang_2016, Zhang_Tian_Kong_Zhong_Fu_2018, Niu_Wen_Ren_Zhang_Yang_Wang_Zhang_Cao_Shen_2020, Wang_Yu_Wu_Gu_Liu_Dong_Qiao_Loy_2019, zhong2024languageguided}, such as deraining \cite{Quan_Yu_Liang_Yang_2021, Qian_Tan_Yang_Su_Liu_2018, Yang_Tan_Feng_Guo_Yan_Liu_2020, Zhu_Fu_Lischinski_Heng_2017}, dehazing \cite{Cai_Xu_Jia_Qing_Tao_2016, Li_Peng_Wang_Xu_Feng_2017, Ren_Ma_Zhang_Pan_Cao_Liu_Yang_2018, Berman_Treibitz_Avidan_2016, Zhang_Patel_2018, Zhang_Ren_Zhang_Zhang_Nie_Xue_Cao_2022}, desnowing \cite{Liu_Jaw_Huang_Hwang_2018, Zhang_Li_Yu_Luo_Li_2021, Jose_Valanarasu_Yasarla_Patel_2022}, and shadow removal~\cite{le2019shadow, ding2019argan, 9710914, 9710813}. However, many image processing methods are trained on images \cite{Li_Wu_Lin_Liu_Zha_2018, Zhang_Sindagi_Patel_2020, Zhang_Li_Yu_Luo_Li_2021, Yasarla_Sindagi_Patel_2020, Quan_Deng_Chen_Ji_2019, Li_Cheong_Tan_2019, Du_Xu_Qiu_Zhen_Zhang_2020} with a single degradation, leading to suboptimal performance when faced with images containing multiple degradation types. Another option is to develop a unified model capable of addressing diverse degradation types simultaneously~\cite{Li_Tan_Cheong_2020, Jose_Valanarasu_Yasarla_Patel_2022, Fan_Chen_Yuan_Hua_Yu_Chen_2018}. However, these methods lack controllability, making it challenging to adhere to user instructions.

In our pursuit of controllability, we begin by focusing on a specific image processing scenario—the blind image decomposition (BID) setting~\cite{han2022bid}. BID demands a framework that can handle multiple types of image degradation concurrently, treating degraded images as arbitrary combinations of diverse, independent components. For example, the deraining process can be seen as the task of dissecting an image with rain components into distinct elements: a rain-free image and a mask indicating the related rainy component.
\begin{figure}[!htb]
\centering
  \includegraphics[width=0.8\linewidth]{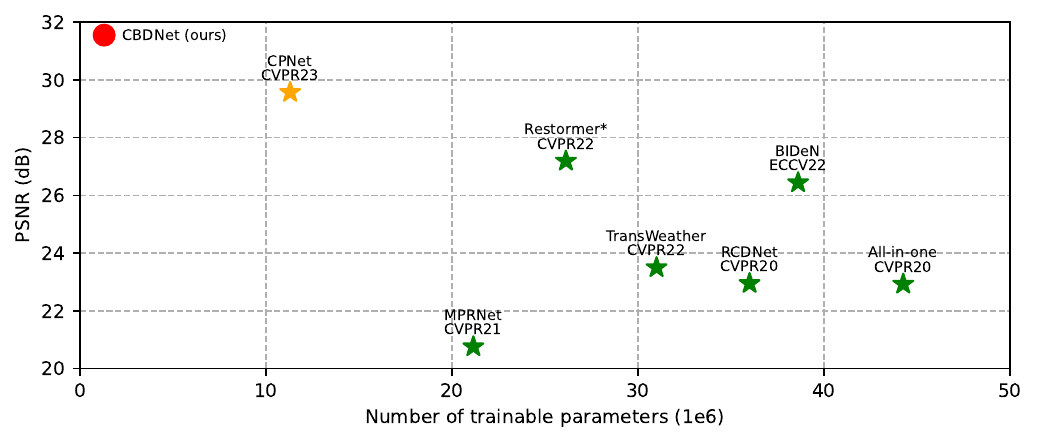}
  \caption{\textbf{Comparison between our proposed method CBDNet and the state-of-the-art methods on the BID task.} The input images have rain streaks, raindrops, snow and haze, and a clean image is the output, which represents the most challenging task in BID. Here, Restormer* denotes the version of Restormer \cite{Zamir_Arora_Khan_Hayat_Khan_Yang_2022} that we trained on the BID dataset with a fixed resolution. Our system not only exhibits a superior image restoration effect but also has a lower number of trainable parameters. }
  \label{fig:pp}
\end{figure}
We propose adding controllability in BID by integrating a user prompt, enabling the selective removal of components as specified by the user. When instructed to separate and reconstruct every image component (rain, fog, watermark, \textit{etc.}), controllable BID is equivalent to standard BID. If the instruction is to only restore the clean image, the task degenerates into an image restoration task \cite{Quan_Yu_Liang_Yang_2021, Pan_Liu_Sun_Zhang_Liu_Ren_Li_Tang_Lu_Tai_2018}. As such, the controllable BID framework is more general and flexible, which makes it better able to meet the complex and varying needs of image processing. Figure~\ref{fig:pipeline} illustrates our workflow. 
The controllable BID module first predicts the specific types of degradation present in input images, which helps users give prompts. After that, it performs the requested image processing.

To enable controllable BID, we propose the controllable blind image decomposition network (CBDNet). This efficient structure utilizes a multi-scale, hierarchical encoder-decoder backbone. User control is achieved through the proposed decomposition, controllability, and recombination blocks, where input feature maps are split into channels and aggregated based on user instructions. These blocks introduce minimal computational overhead while effectively following user intentions. Besides controllability, our system surpasses existing methods in both efficiency and accuracy in BID tasks. In Figure~\ref{fig:pp}, we compare the performance of CBDNet against existing methods on the BID task.

To further study controllable BID in real-world applications, we construct a more comprehensive and challenging multi-domain degradation removal dataset. This dataset includes a variety of degradation components from different domains: from the weather domain, we have rain streaks \cite{Yang_Tan_Feng_Liu_Guo_Yan_2017}, raindrops \cite{raindrop}, snow \cite{Liu_Jaw_Huang_Hwang_2018}, and haze \cite{Sakaridis_Dai_Van_Gool_2018}; from the lighting domain, we incorporate flare \cite{Wu_He_Xue_Garg_Chen_Veeraraghavan_Barron_2021}, reflections \cite{Zhang_Ng_Chen_2018}, and shadows \cite{Wang_Li_Yang_2018, Cun_Pun_Shi_2020}; and from the obstruction domain, we feature fence \cite{Liu_Lai_Yang_Chuang_Huang_2020} and watermark \cite{Liu_Zhu_Bai_2021}. Our contributions are summarized below.
\begin{itemize}
  \item We bring controllability to the BID task, which, based on a user prompt, selectively removes specific components from an image. The inclusion of user prompts in BID 
  brings its image processing capability closer to the complex and varying needs of users in real-world scenarios.
  
  \item  For controllable BID, we construct a multi-domain degradation removal dataset, comprised of nine common types of degradation, to support future research in this field. 
  
  \item We propose an efficient and effective architecture, controllable blind image decomposition network (CBDNet). Experimentally, our method achieves state-of-the-art performance in BID tasks as well as controllable blind image decomposition aligning with user prompts.
\end{itemize}

\section{Related work}
\textbf{Multi-degradation removal} has drawn increasing interest in the community \cite{Pan_Liu_Sun_Zhang_Liu_Ren_Li_Tang_Lu_Tai_2018, Zamir_Arora_Khan_Hayat_Khan_Yang_Shao_2021, Zamir_Arora_Khan_Hayat_Khan_Yang_2022, Wang_Cun_Bao_Zhou_Liu_Li_2022, Liang_Cao_Sun_Zhang_Van_Gool_Timofte_2021, Zhang_Huang_Yao_Yang_Yu_Zhou_Zhao, 10204072}. These approaches generally attempt to deploy a unified model framework to tackle various types of degradation but necessitate task-specific pre-trained parameters, which might be less useful when there are more degradations. For example, all-in-one \cite{Li_Tan_Cheong_2020} addresses multiple weather removal by using a single set of pre-trained parameters, but its effectiveness decreases with the increase in the number of types of degradation due to the need of multiple encoders. Subsequent approaches~\cite{Potlapalli_Zamir_Khan_Khan_2023, Wang_Yu_Zhang_2022, Li_Liu_Hu_Wu_Lv_Peng, Zhu_Wang_Fu_Yang_Guo_Dai_Qiao_Hu, chen2022learning, ma2023prores, li2023promptinprompt, 10204770, yang2023languagedriven} adopt a single encoder and decoder to effectively handle various types of degradation. However, these methods are typically designed to address a single degradation at a time. In this study, we introduce a unified model designed to address various tasks with controllability.

\noindent\textbf{Blind image decomposition}
(BID)~\cite{han2022bid} solves every image degradation task by treating an image as a combination of degradation layers over clean image layers, drawing inspiration from the blind source separation problem \cite{Kun_Gai_Zhenwei_Shi_Changshui_Zhang_2008, Cichocki_Amari_2002, Gu_Meng_Zuo_Zhang_2017, Hyvärinen_Oja_2000}. BID perceives degraded images as a combination of several independent components, thus offering a unified framework for handling various types of image degradation at once.
BIDeN~\cite{han2022bid}, which requires an additional pair of generators and discriminators for each new type of degradation supported, suffers from efficiency issues. CPNet \cite{Wang_2023_CVPR} is a single encoder-decoder structure, but it exhibits limitations in accurately reconstructing degradation components. These models also lack flexibility and controllability, and we aim to fill this gap.
\begin{figure}[!htbp]
\includegraphics[width=\linewidth]{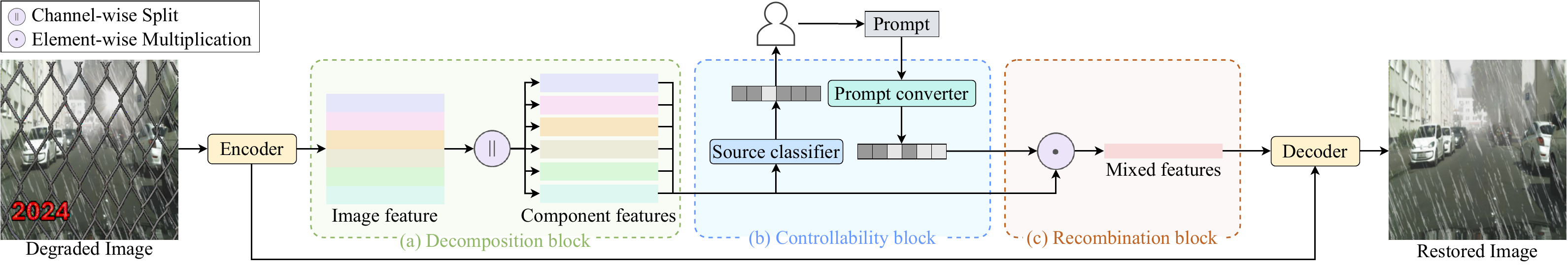}
\caption{\textbf{Architecture of CBDNet.} Upon receiving an input image with various degradation, the encoder transforms it into a deep feature map. This map is then split into several component feature maps by the decomposition block. Each component feature corresponds to a type of degradation. Within the controllablility block, the source classifier utilizes these component feature maps, facilitating source classification and enabling the user to give an instruction prompt easily. This prompt is subsequently converted by the prompt converter into a categorical vector indicating the image processing to be executed. In the recombination block, the chosen component feature maps are mixed with the categorical vector of the prompt and then reconstructed into a restored image by the decoder.
}
\vspace{-1em}
\label{fig:net}
\end{figure}

\noindent\textbf{Controllable image processing}
Recently, user controllability has been widely explored in visual generation tasks \cite{zhang2023adding,avrahami2022blended, brooks2022instructpix2pix,
		gafni2022make,hertz2022prompt, kawar2022imagic,
		ramesh2022hierarchical, brooks2023instructpix2pix, bai2023textir}.
PIP \cite{hertz2022prompt} proposed a prompt-to-prompt editing network to provide an image editing interface using only text prompts. In their pioneering work, \cite{zhang2023adding} introduced ControlNet, an advanced adaptation of a pre-trained text-to-image diffusion model. Some latest works \cite{kirillov2023segment} and \cite{zou2023segment} have led to the development of prompt-controlled general image segmentation models.
Drawing inspiration from these works, we introduce controllability to BID. Our method is designed to selectively remove elements from an image based on specific user instructions. Our method simplifies the process for users, requiring no professional knowledge. It initially presents all source components occurring within an image for easy reference. Users can then conveniently choose to either remove or retain components.

\section{Method}

\subsection{Problem formulation}

Let $D = \{D_1, D_2, \ldots, D_N\}$ be a set of source image domains with each domain $D_i$ containing a set of components $c_i$. We define a set of distinct indices $J = \{j_1, j_2, \ldots, j_m\}$ such that $j_i \in \{1, 2, \ldots, N\}$, $j_i \neq j_k$ for all $i \neq k$ and $1\leq m\leq N$, ensuring each $c_i$ comes from a uniquely specified domain.
For a composite function $f$, input $x$ is calculated by $ x = f(\{c_{j_i}\}_{i=1}^m)$, where $c_{j_i} \in D_{j_i}$ corresponds to an component from the source domain $D_{j_i}$. Given a proper prompt $P$, any subset of $J$ can be selected, denoted as $S = \{s_1, s_2, \ldots, s_l\}$, $1\leq l\leq m$ and $S\in J$.
Controllable BID aims to find function $g$ to decompose $x$ according to $S$. Output $y'=g(x,S)$ is expected to close to $y = f(\{c_{s_i}\}_{i=1}^l)$, where $y$ is the output of $f$ for the elements indexed by $S$; $x_{s_i}$ is the element chosen from $D_{s_i}$.

\subsection{Overall structure}

Taking inspiration from the recent success of the image restoration network Restormer~\cite{Zamir_Arora_Khan_Hayat_Khan_Yang_2022}, we first establish a strong baseline within the BID framework. Our baseline adopts the transformer-based encoder-decoder architecture design of Restormer, with specific adjustments made to adapt it to the BID setting. We refer to this adapted version of Restormer as Restormer*.

To add controllability, the proposed CBDNet extends this baseline by integrating three critical modules: the decomposition block, the controllability block, and the recombination block, as shown in Figure~\ref{fig:net}. The encoder transforms the input image $ I\in \mathcal{R}^{H\times W\times 3} $ into feature map $ F_d \in \mathcal{R}^{H\times W\times C_d}$, with $H\times W$ representing the resolution of $ I$ and $C_d$ denoting the output channels of encoder. $ F_d $ is then split into a set of component feature maps $\{F_{c_{i}}\}_{i=1}^N$ in the decomposition block, where $ F_{c_{i}}\in \mathcal{R}^{H\times W\times \frac{C_d}{N}} $ and $N$ is the max number of source domains. The controllability block uses $\{F_{c_{i}}\}_{i=1}^N$ to obtain the source prediction $ V_s\in \mathcal{R}^{N} $ which indicates the existence of each component. 
Leveraging $V_s$, user can provide an instruction prompt $P$ about the anticipated processing outcome, which will be converted to a vector $ V_c \in \mathcal{R}^{N} $. 
Here, $ V_s $ and $ V_c $ are categorical vectors, where each element corresponds to a specific component, with $1$ indicating either existence or retention of that component, and $0$ standing for either non-existence or removal of that component. $ V_c $ and $\{F_{c_{i}}\}_{i=1}^N$ are employed to obtain the mixed feature map $ F_r \in \mathcal{R}^{H\times W\times \frac{C}{N}}$ in the recombination block. Finally, the decoder processes $ F_r $, generating output $ I' \in \mathcal{R}^{H\times W\times 3} $.

\subsection{Decomposition block}
\begin{figure}[t]
    \includegraphics[width=\linewidth]{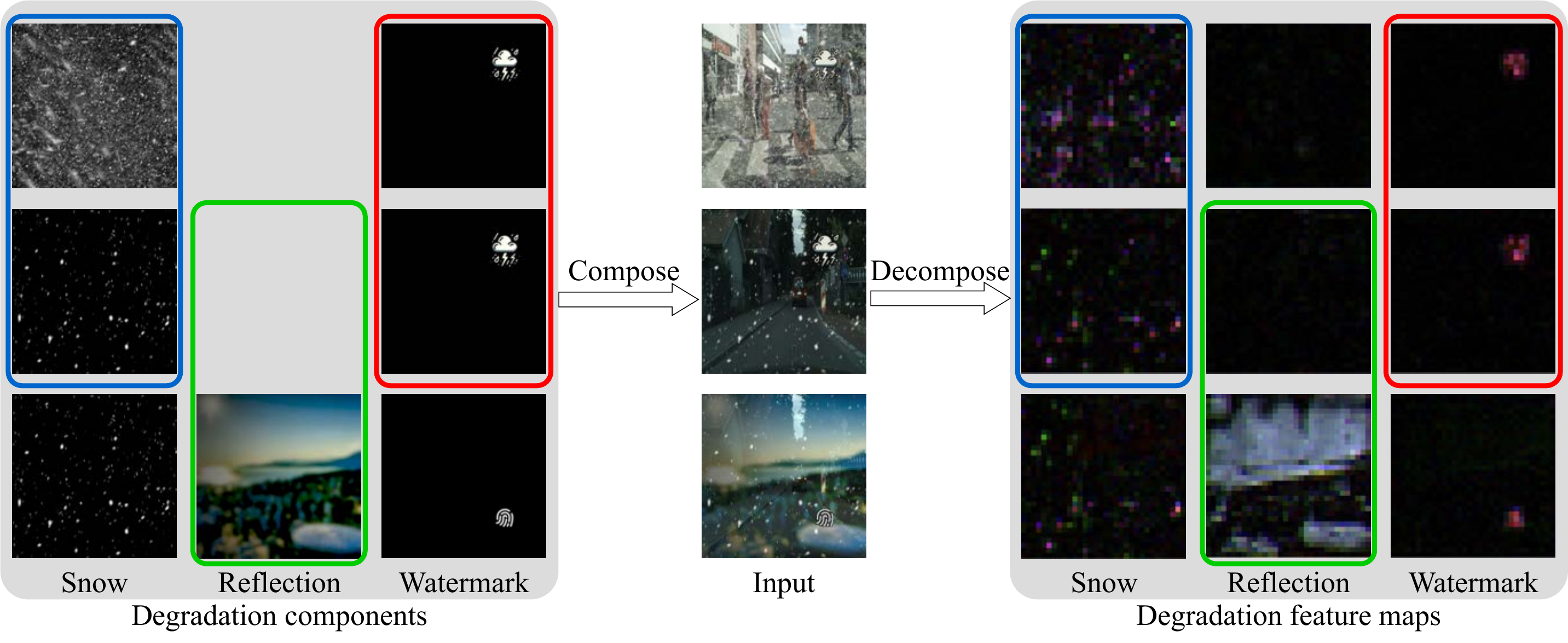}
    \caption{\textbf{Visualization of feature maps output by the decomposition block.} The input image is composed of corresponding degradations on the left side, and then produces the feature maps on the right side after passing through the encoder and decomposition block. The red boxes highlight the near-identical nature of feature maps for the same component across different images. In contrast, the blue boxes reveal complete dissimilarities in feature maps for different components. Furthermore, the green boxes illustrate that absent degradations in the input image lead to feature maps without significant patterns, distinct from those with identifiable degradations.}
    \label{fig:visual}
\end{figure}

We propose that the feature map produced by the encoder already contains characteristics of all components, though they are intermixed in the channel dimension. As long as the mixed features can be segregated according to their respective components, we can achieve separation of different components without increasing the number of learnable parameters. Building upon this assumption, we introduce a simple decomposition block capable of separating features of individual components, as illustrated in Figure~\ref{fig:net} (a). This design essentially performs a split operation along the channel dimension, dividing the encoder's output feature map $F_d$ into multiple sub-feature maps $\{F_{c_{i}}\}_{i=1}^N$. By having each sub-feature map $F_{c_{i}}$ reconstructed by the decoder and supervised by the loss against the ground truth, we impart explicit meaning to each channel of the encoder's output. In Figure~\ref{fig:visual}, we exhibit a visualization of the feature maps generated by the decomposition block, indicating that the decomposition block can effectively separate various degradations, and the generated feature maps can uniquely represent the corresponding degradation components.

\subsection{Controllability block}
\label{sec:prompt}

The controllability block comprises the source classifier and the prompt converter, as shown in Figure~\ref{fig:net}. We leverage source classifier to predict the source components involved in mixing. We then translate the source prediction into a description of the input image for the user. After that, the user can specify whether to remove each component by giving a text prompt. If the user does not input any instructions, the default prompt will guide the model to remove all degradation components, at which point the controllable blind image decomposition task degenerates into a classical image restoration task.

\noindent\textbf{Source classifier.}
We employ a tiny CNN to predict the source component categorical vector $V_s$ based on the output $\{F_{c_{i}}\}_{i=1}^N$ from the decomposition block. This CNN network is composed of two convolutional layers, two pooling layers, and a fully connected layer.

\noindent\textbf{Prompt converter.}
We use a frozen BERT encoder \cite{devlin2019bert} to transform the input prompt $P$ into a text embedding, then employ a lightweight attention module to convert it into the categorical vector $V_c$. 

\subsection{Recombination block}
We introduce an effective recombination block, as illustrated in Figure~\ref{fig:net}.
The pivotal operation within the recombination block is the multiplication of the feature maps with the categorical vector $V_c$, which serves to blend the selected components. The prompt classification vector is of length equivalent to the maximum number of source domains, representing the demands for each component. The recombination block is also parameter-free.

\subsection{Loss Functions}
\label{sec:loss}
During the training phase, we require CBDNet to reconstruct all components $\{y'_{i}\}_{i=1}^N$, as well as an mixed image \(y'_{N+1}\)that randomly removes a selection of these components, where \(N\) is the max number of source domains. To enhance the texture and perceptual quality of results, CBDNet uses a combination of smooth L1 loss, VGG loss \cite{Simonyan_Zisserman_2015}, LPIPS loss \cite{Zhang_Isola_Efros_Shechtman_Wang_2018}, and BCE loss, which promotes the generation of well separated and realistic images.

The texture loss \(\mathcal{L}_{\text{texture}}\) is written as:
\begin{equation}
\mathcal{L}_{\text{texture}} = \lambda_{\text{smoothL1}}\sum_{i=1}^{N+1} \mathcal{L}_{\text{smoothL1}}(y'_i, y_i).
\end{equation}

The perceptual loss \(\mathcal{L}_{\text{perceptual}}\) is calculated as:
\begin{equation}
\begin{aligned}
\mathcal{L}_{\text{perceptual}} = \, & \lambda_{\text{VGG}} \left( \mathcal{L}_{\text{VGG}}(y'_1, y_1) + \mathcal{L}_{\text{VGG}}(y'_{N+1}, y_{N+1}) \right) \\
& + \lambda_{\text{LPIPS}} \left( \mathcal{L}_{\text{LPIPS}}(y'_1, y_1) + \mathcal{L}_{\text{LPIPS}}(y'_{N+1}, y_{N+1}) \right).
\end{aligned}
\end{equation}

\begin{figure*}[t]
  \begin{minipage}[t]{0.135\linewidth} 
    \centering 
    \includegraphics[width=1\linewidth, height=0.5\linewidth]{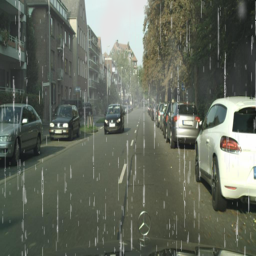}
  \end{minipage} 
    \begin{minipage}[t]{0.135\linewidth} 
    \centering 
        \includegraphics[width=1\linewidth, height=0.5\linewidth]{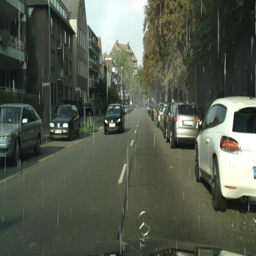}
  \end{minipage}
      \begin{minipage}[t]{0.135\linewidth} 
    \centering 
        \includegraphics[width=1\linewidth, height=0.5\linewidth]{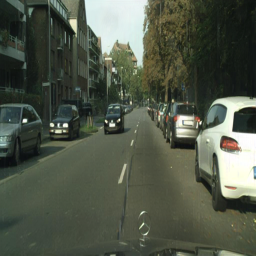}
  \end{minipage}
      \begin{minipage}[t]{0.135\linewidth} 
    \centering 
        \includegraphics[width=1\linewidth, height=0.5\linewidth]{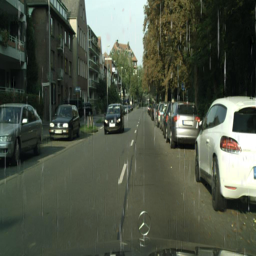}
  \end{minipage}  
   \begin{minipage}[t]{0.135\linewidth} 
    \centering 
    \includegraphics[width=1\linewidth, height=0.5\linewidth]{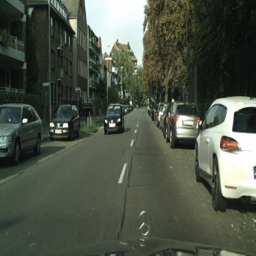}
  \end{minipage} 
  \begin{minipage}[t]{0.135\linewidth} 
    \centering 
    \includegraphics[width=1\linewidth, height=0.5\linewidth]{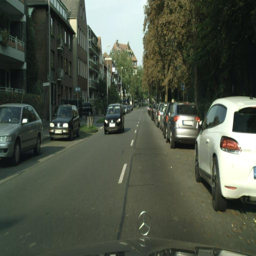}
  \end{minipage} 
    \begin{minipage}[t]{0.135\linewidth} 
    \centering 
        \includegraphics[width=1\linewidth, height=0.5\linewidth]{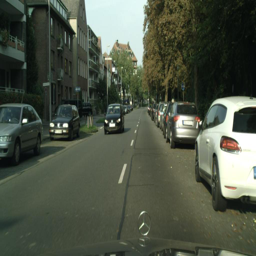}
  \end{minipage} 
  \begin{minipage}[t]{0.135\linewidth} 
    \centering 
    \includegraphics[width=1\linewidth, height=0.5\linewidth]{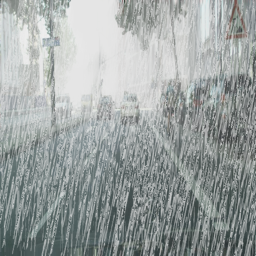}
  \end{minipage} 
    \begin{minipage}[t]{0.135\linewidth} 
    \centering 
        \includegraphics[width=1\linewidth, height=0.5\linewidth]{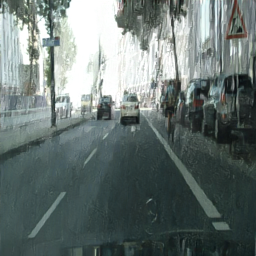}
  \end{minipage}
      \begin{minipage}[t]{0.135\linewidth} 
    \centering 
        \includegraphics[width=1\linewidth, height=0.5\linewidth]{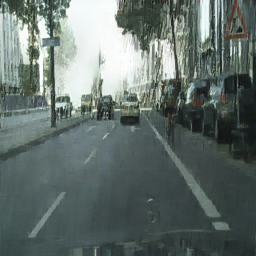}
  \end{minipage}
      \begin{minipage}[t]{0.135\linewidth} 
    \centering 
        \includegraphics[width=1\linewidth, height=0.5\linewidth]{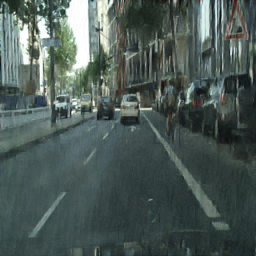}
  \end{minipage}  
   \begin{minipage}[t]{0.135\linewidth} 
    \centering 
    \includegraphics[width=1\linewidth, height=0.5\linewidth]{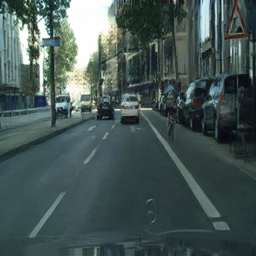}
  \end{minipage} 
  \begin{minipage}[t]{0.135\linewidth} 
    \centering 
    \includegraphics[width=1\linewidth, height=0.5\linewidth]{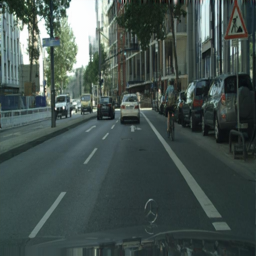}
  \end{minipage} 
    \begin{minipage}[t]{0.135\linewidth} 
    \centering 
        \includegraphics[width=1\linewidth, height=0.5\linewidth]{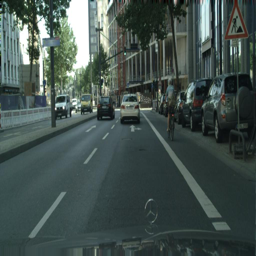}
  \end{minipage} 
  \begin{minipage}[t]{0.135\linewidth} 
    \centering 
    \includegraphics[width=1\linewidth, height=0.5\linewidth]{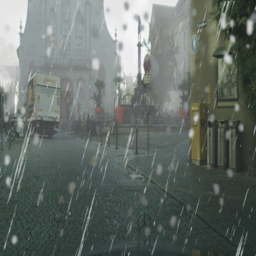}
  \end{minipage} 
    \begin{minipage}[t]{0.135\linewidth} 
    \centering 
        \includegraphics[width=1\linewidth, height=0.5\linewidth]{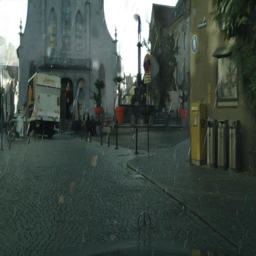}
  \end{minipage}
      \begin{minipage}[t]{0.135\linewidth} 
    \centering 
        \includegraphics[width=1\linewidth, height=0.5\linewidth]{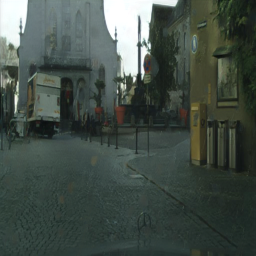}
  \end{minipage}
      \begin{minipage}[t]{0.135\linewidth} 
    \centering 
        \includegraphics[width=1\linewidth, height=0.5\linewidth]{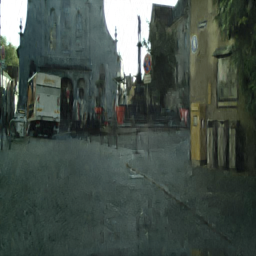}
  \end{minipage}  
   \begin{minipage}[t]{0.135\linewidth} 
    \centering 
    \includegraphics[width=1\linewidth, height=0.5\linewidth]{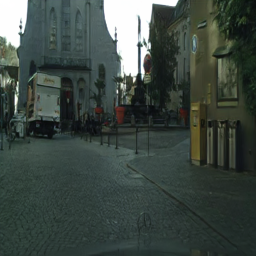}
  \end{minipage} 
  \begin{minipage}[t]{0.135\linewidth} 
    \centering 
    \includegraphics[width=1\linewidth, height=0.5\linewidth]{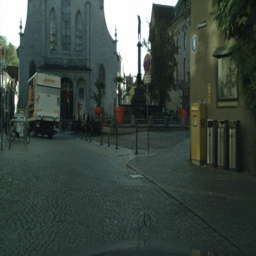}
  \end{minipage} 
    \begin{minipage}[t]{0.135\linewidth} 
    \centering 
        \includegraphics[width=1\linewidth, height=0.5\linewidth]{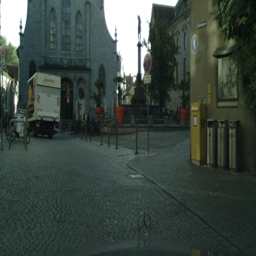}
  \end{minipage} 
  \begin{minipage}[t]{0.135\linewidth} 
    \centering 
    \includegraphics[width=1\linewidth, height=0.5\linewidth]{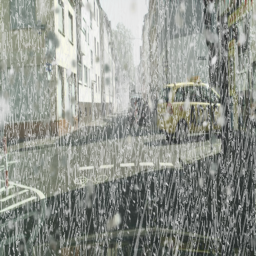}
  \end{minipage} 
    \begin{minipage}[t]{0.135\linewidth} 
    \centering 
        \includegraphics[width=1\linewidth, height=0.5\linewidth]{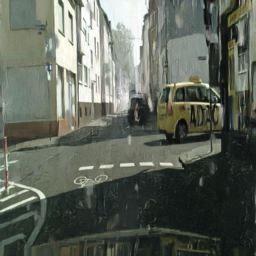}
  \end{minipage}
      \begin{minipage}[t]{0.135\linewidth} 
    \centering 
        \includegraphics[width=1\linewidth, height=0.5\linewidth]{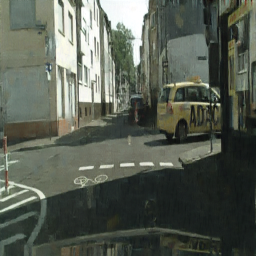}
  \end{minipage}
      \begin{minipage}[t]{0.135\linewidth} 
    \centering 
        \includegraphics[width=1\linewidth, height=0.5\linewidth]{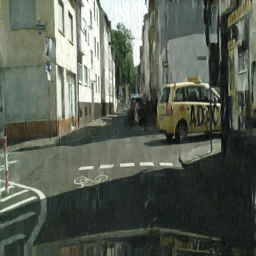}
  \end{minipage}  
   \begin{minipage}[t]{0.135\linewidth} 
    \centering 
    \includegraphics[width=1\linewidth, height=0.5\linewidth]{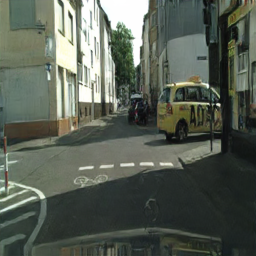}
  \end{minipage} 
  \begin{minipage}[t]{0.135\linewidth} 
    \centering 
    \includegraphics[width=1\linewidth, height=0.5\linewidth]{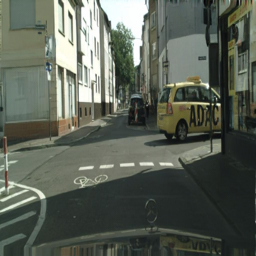}
  \end{minipage} 
    \begin{minipage}[t]{0.135\linewidth} 
    \centering 
        \includegraphics[width=1\linewidth, height=0.5\linewidth]{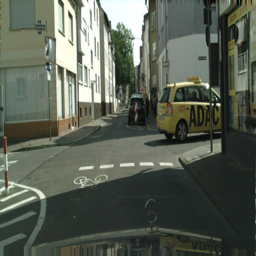}
  \end{minipage} 
    \begin{minipage}[t]{0.134\linewidth} 
    \centering 
    \text{\scriptsize Input}
  \end{minipage} 
    \begin{minipage}[t]{0.136\linewidth} 
    \centering 
    \text{\scriptsize MPRNet}
  \end{minipage}
      \begin{minipage}[t]{0.140\linewidth} 
    \centering 
    \text{\scriptsize All-in-one}
  \end{minipage}
      \begin{minipage}[t]{0.136\linewidth} 
    \centering 
    \text{\scriptsize Restormer}
  \end{minipage} 
    \begin{minipage}[t]{0.136\linewidth} 
    \centering 
      \text{\scriptsize BIDeN}
  \end{minipage}  
    \begin{minipage}[t]{0.136\linewidth} 
    \centering 
      \text{\scriptsize Ours}
  \end{minipage} 
    \begin{minipage}[t]{0.136\linewidth} 
    \centering 
        \text{\scriptsize GT}
  \end{minipage}
  \caption{\textbf{Qualitative results of Task I: Multi-weather removal in driving.} 
  Row 1-4 represents the Case (3)-(6) in Table~\ref{tab:task1-a}. Our proposed CBDNet demonstrates good performance in the removal of all components. It produces artifact-free results that are aesthetically appealing, whereas all baseline methods tend to leave behind some components, particularly in row 2 and row 4.}
  \label{fig:task1-a-fig}
  \vspace{-0.5em}
\end{figure*}
\begin{table}[!htb]
\centering
\fontsize{7}{3}\selectfont

\setlength{\tabcolsep}{1pt}
\caption{\textbf{Quantitative results on Task I (Multi-weather removal in driving).} The performance of SSIM and PSNR has been assessed in 6 different cases, which are: (1) rain streak, (2) rain streak + snow, (3) rain streak + light haze, (4) rain streak + heavy haze, (5) rain streak + moderate haze + raindrop, and (6) rain streak + snow + moderate haze + raindrop. Restormer* is a reimplementation of Restormer. Our proposed methods achieves the best performance across all cases.}
\begin{tabular}{@{}lcc|cc|cc|cc|cc|cc@{}}
\toprule
Case&\multicolumn{2}{c|}{MPRNet}&\multicolumn{2}{c|}{All-in-one}&\multicolumn{2}{c|}{Restormer*}&\multicolumn{2}{c|}{BIDeN}&\multicolumn{2}{c|}{CPNet} &\multicolumn{2}{c}{Ours}\cr
&PSNR$\uparrow$ & SSIM$\uparrow$ & PSNR$\uparrow$ & SSIM$\uparrow$ & PSNR$\uparrow$ & SSIM$\uparrow$ & PSNR$\uparrow$ & SSIM$\uparrow$ & PSNR$\uparrow$ & SSIM$\uparrow$ & PSNR$\uparrow$ & SSIM$\uparrow$ \cr

\midrule
\multicolumn{1}{c}{(1)}&33.39 &0.945 &32.38 &0.937 &\underline{37.87} &\underline{0.972} &30.89 &0.932 &33.95 &0.948 &\textbf{42.08} &\textbf{0.993} \cr
\multicolumn{1}{c}{(2)}&30.52 &0.909 &28.45 &0.892 &32.68 &0.919 &29.34 &0.899 &\underline{33.42} &\underline{0.937} &\textbf{37.01} &\textbf{0.980} \cr
\multicolumn{1}{c}{(3)}&23.98 &0.900 &27.14 &0.911 &32.15 &\underline{0.954} &28.62 &0.919 &\underline{32.99} &0.932 &\textbf{36.00} &\textbf{0.985} \cr
\multicolumn{1}{c}{(4)}&18.54 &0.829 &19.67 &0.865 &28.74 &\underline{0.924} &26.77 &0.891 &\underline{29.02} &0.908 &\textbf{33.02} &\textbf{0.974} \cr
\multicolumn{1}{c}{(5)}&21.18 &0.846 &24.23 &0.889 &28.03 &0.903 &27.11 &0.898 &\underline{30.07} &\underline{0.925} &\textbf{32.64} &\textbf{0.967} \cr
\multicolumn{1}{c}{(6)}&20.76 &0.812 &22.93 &0.846 &27.19 &0.863 &26.44 &0.870 &\underline{29.57} &\underline{0.914} &\textbf{31.55} &\textbf{0.952} \cr
\bottomrule
\end{tabular}

\label{tab:task1-a}
\end{table}

The overall loss function is:
\begin{equation}
\mathcal{L}_{\text{total}} = \mathcal{L}_{\text{texture}} + \mathcal{L}_{\text{perceptual}} + \mathcal{L}_{\text{BCE}}.
\end{equation}

\section{Experiment}

\subsection{Tasks and Datasets}
We conduct analysis on two BID tasks: Task I, Multi-weather removal in driving; Task II, Real-world bad weather removal. To better emulate the intricacies of real-world scenarios, we further construct a exceptionally challenging Task III, Multi-degradation removal. This dataset includes nine degradation components from three domains: rain streaks \cite{Yang_Tan_Feng_Liu_Guo_Yan_2017}, raindrops \cite{raindrop}, snow \cite{Liu_Jaw_Huang_Hwang_2018}, and haze \cite{Sakaridis_Dai_Van_Gool_2018} from the weather domain; flare \cite{Wu_He_Xue_Garg_Chen_Veeraraghavan_Barron_2021}, reflections \cite{Zhang_Ng_Chen_2018}, and shadows \cite{Wang_Li_Yang_2018, Cun_Pun_Shi_2020} from the lighting domain; lastly, fence \cite{Liu_Lai_Yang_Chuang_Huang_2020} and watermark \cite{Liu_Zhu_Bai_2021} from the domain of obstructions. For details of datasets, evaluation metrics, implementation, and training, please refer to the supplementary materials.

\begin{figure*}[!htb]
	\begin{minipage}{1\textwidth}
		\begin{minipage}[c]{0.07\linewidth}
			\hfill 
			\text{\scriptsize Input}
		\end{minipage}%
		\begin{minipage}[c]{0.115\linewidth}
			\includegraphics[width=0.98\linewidth, height=0.98\linewidth]{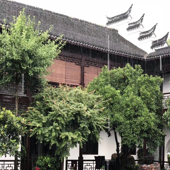}
		\end{minipage}%
		\begin{minipage}[c]{0.115\linewidth}
			\includegraphics[width=0.98\linewidth, height=0.98\linewidth]{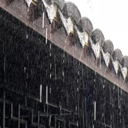}
		\end{minipage}%
		\begin{minipage}[c]{0.115\linewidth}
			\includegraphics[width=0.98\linewidth, height=0.98\linewidth]{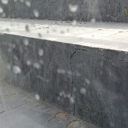}
		\end{minipage}%
		\begin{minipage}[c]{0.115\linewidth}
			\includegraphics[width=0.98\linewidth, height=0.98\linewidth]{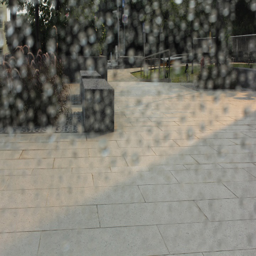}
		\end{minipage}%
		\begin{minipage}[c]{0.115\linewidth}
			\includegraphics[width=0.98\linewidth, height=0.98\linewidth]{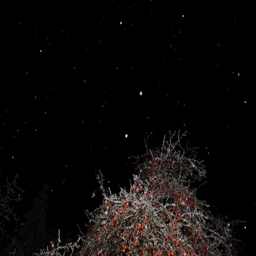}
		\end{minipage}%
		\begin{minipage}[c]{0.115\linewidth}
			\includegraphics[width=0.98\linewidth, height=0.98\linewidth]{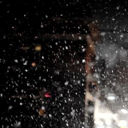}
		\end{minipage}%
		\begin{minipage}[c]{0.115\linewidth}
			\includegraphics[width=0.98\linewidth, height=0.98\linewidth]{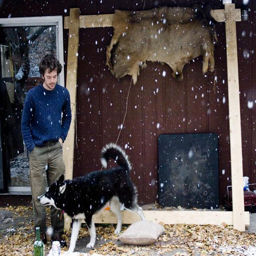}
		\end{minipage}%
		\begin{minipage}[c]{0.115\linewidth}
			\includegraphics[width=0.98\linewidth, height=0.98\linewidth]{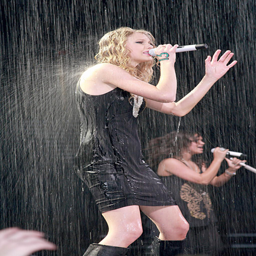}
		\end{minipage}%

		\begin{minipage}[c]{0.07\linewidth}
			\hfill 
			\text{\scriptsize BIDeN}
		\end{minipage}%
		\begin{minipage}[c]{0.115\linewidth}
			\includegraphics[width=0.98\linewidth, height=0.98\linewidth]{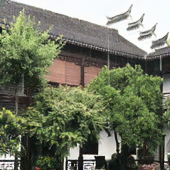}
		\end{minipage}%
		\begin{minipage}[c]{0.115\linewidth}
			\includegraphics[width=0.98\linewidth, height=0.98\linewidth]{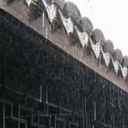}
		\end{minipage}%
		\begin{minipage}[c]{0.115\linewidth}
			\includegraphics[width=0.98\linewidth, height=0.98\linewidth]{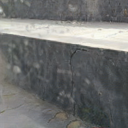}
		\end{minipage}%
		\begin{minipage}[c]{0.115\linewidth}
			\includegraphics[width=0.98\linewidth, height=0.98\linewidth]{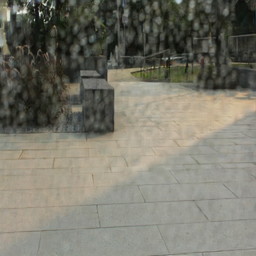}
		\end{minipage}%
		\begin{minipage}[c]{0.115\linewidth}
			\includegraphics[width=0.98\linewidth, height=0.98\linewidth]{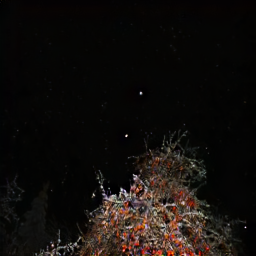}
		\end{minipage}%
		\begin{minipage}[c]{0.115\linewidth}
			\includegraphics[width=0.98\linewidth, height=0.98\linewidth]{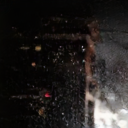}
		\end{minipage}%
		\begin{minipage}[c]{0.115\linewidth}
			\includegraphics[width=0.98\linewidth, height=0.98\linewidth]{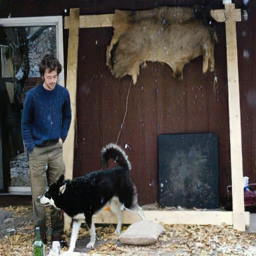}
		\end{minipage}%
		\begin{minipage}[c]{0.115\linewidth}
			\includegraphics[width=0.98\linewidth, height=0.98\linewidth]{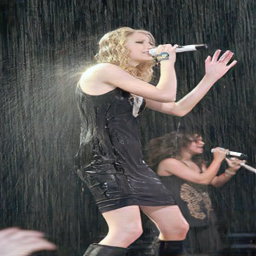}
		\end{minipage}%

		\begin{minipage}[c]{0.07\linewidth}
			\hfill 
			\text{\scriptsize Ours}
		\end{minipage}%
		\begin{minipage}[c]{0.115\linewidth}
			\includegraphics[width=0.98\linewidth, height=0.98\linewidth]{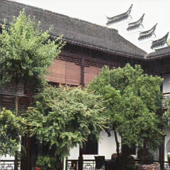}
            \centering\text{\scriptsize Case (1)}
		\end{minipage}%
		\begin{minipage}[c]{0.115\linewidth}
			\includegraphics[width=0.98\linewidth, height=0.98\linewidth]{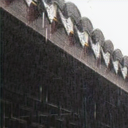}
               \centering\text{\scriptsize Case (2)}
		\end{minipage}%
		\begin{minipage}[c]{0.115\linewidth}
			\includegraphics[width=0.98\linewidth, height=0.98\linewidth]{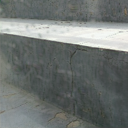}
               \centering\text{\scriptsize Case (3)}
		\end{minipage}%
		\begin{minipage}[c]{0.115\linewidth}
			\includegraphics[width=0.98\linewidth, height=0.98\linewidth]{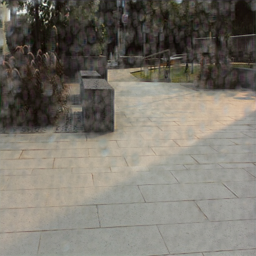}
               \centering\text{\scriptsize Case (4)}
		\end{minipage}%
		\begin{minipage}[c]{0.115\linewidth}
			\includegraphics[width=0.98\linewidth, height=0.98\linewidth]{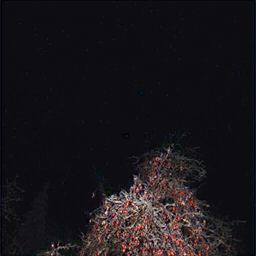}
               \centering\text{\scriptsize Case (5)}
		\end{minipage}%
		\begin{minipage}[c]{0.115\linewidth}
			\includegraphics[width=0.98\linewidth, height=0.98\linewidth]{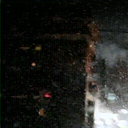}
               \centering\text{\scriptsize Case (6)}
		\end{minipage}%
		\begin{minipage}[c]{0.115\linewidth}
			\includegraphics[width=0.98\linewidth, height=0.98\linewidth]{figures/task2/I7_00149.png}
               \centering\text{\scriptsize Case (7)}
		\end{minipage}%
		\begin{minipage}[c]{0.115\linewidth}
			\includegraphics[width=0.98\linewidth, height=0.98\linewidth]{figures/task2/I8_02168.png}
               \centering\text{\scriptsize Case (8)}
		\end{minipage}%
	\end{minipage}%

\caption{Qualitative results of Task II: Real-world bad
weather removal. We showcase the performance of methods
across three real-world nature datasets, each encompassing distinct weather conditions: rain streak, raindrop, and snow. }
\label{fig:task2-fig}
\end{figure*}

\begin{table}[htb]
\centering
\fontsize{7}{3}\selectfont
\setlength{\tabcolsep}{6pt}
\caption{\textbf{Quantitative results of Task II: Real-world bad weather removal.} We evaluate the performance of methods across three real-world nature datasets, each encompassing distinct weather conditions: rain streak, raindrop, and snow. In most cases, CBDNet has achieved outstanding performance.}
\begin{tabular}{lc|c|c|c|c|c}
\toprule
 \Rows{Method}& \multicolumn{2}{c|}{Rain Streak} & \multicolumn{2}{c|}{Raindrop} & \multicolumn{2}{c}{Snow}\cr
&NIQE$\downarrow$ & BRISQUE$\downarrow$ & NIQE$\downarrow$ & BRISQUE$\downarrow$& NIQE$\downarrow$ & BRISQUE$\downarrow$\cr
\midrule
\multicolumn{1}{c}{Input}& 4.87 & 27.82 & 5.63 & 24.88  & 4.75 &22.68 \cr
\multicolumn{1}{c}{MPRNet}& \textbf{4.10} & 28.66 & 4.87 &29.17  & 4.48 & 25.78 \cr
\multicolumn{1}{c}{BIDeN} & 4.33 & 26.57 & 4.72 & 21.22 & 4.31 & 22.42 \cr  
\multicolumn{1}{c}{CPNet} & 4.13 & 25.58 & \textbf{4.50} & 20.16 & \textbf{4.16} & 21.88 \cr  
\multicolumn{1}{c}{Ours} & 4.31 & \textbf{23.40} & 4.80 & \textbf{19.64} & 4.54 & \textbf{19.31} \cr  
\bottomrule
\end{tabular}
\label{tab:task2}
\end{table}

\subsection{CBDNet is a strong blind image decomposer} 
We compare CBDNet with current state-of-the-art in BID tasks.

\noindent\textbf{Task I: Multi-weather removal in driving.}
This task primarily focuses on the joint removal of rain streak/snow/haze/raindrop application. All methods compared in the experiment are trained using the same BID setting. Quantitative and qualitative results are presented in Table~\ref{tab:task1-a} and Figure~\ref{fig:task1-a-fig}. MPRNet \cite{Zamir_Arora_Khan_Hayat_Khan_Yang_Shao_2021} exhibits commendable performance in simpler scenarios, namely case (1) and case (2), but its effectiveness rapidly drops in more challenging cases. BIDeN \cite{han2022bid} consistently shows strong performance in both simple and complex settings. Leveraging the strong feature extraction and generalization capabilities of its transformer and U-Net architecture, Restormer* excels particularly in simpler scenarios. However, in more demanding settings, Restormer falls short compared to CPNet \cite{Wang_2023_CVPR}, which has undergone extensive self-supervised training on ImageNet \cite{Russakovsky_Deng_Su_Krause_Satheesh_Ma_Huang_Karpathy_Khosla_Bernstein_et}. Overall, our method achieves the best performance across all scenarios.
We also compare the performance of BIDeN and our method on mask reconstruction, with results presented in the supplemental materials.

\begin{figure*}[htb!]
    \begin{minipage}{1\linewidth}
    \centering
        \begin{minipage}[c]{0.1\linewidth}
            \hfill 
            \text{\scriptsize Input\;}
        \end{minipage}%
        \begin{minipage}[c]{0.15\linewidth}
            \includegraphics[width=0.96\linewidth, height=0.48\linewidth]{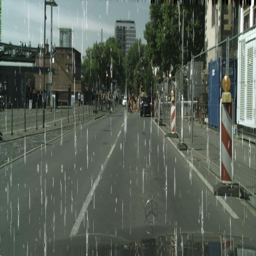}
        \end{minipage}%
        \begin{minipage}[c]{0.15\linewidth}
            \includegraphics[width=0.96\linewidth, height=0.48\linewidth]{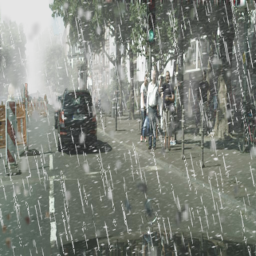}
        \end{minipage}%
        \begin{minipage}[c]{0.15\linewidth}
            \includegraphics[width=0.96\linewidth, height=0.48\linewidth]{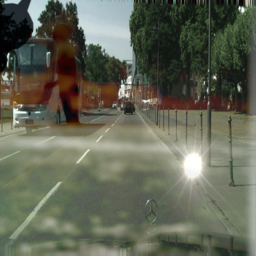}
        \end{minipage}%
        \begin{minipage}[c]{0.15\linewidth}
            \includegraphics[width=0.96\linewidth, height=0.48\linewidth]{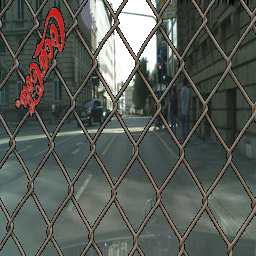}
        \end{minipage}%
        \begin{minipage}[c]{0.15\linewidth}
            \includegraphics[width=0.96\linewidth, height=0.48\linewidth]{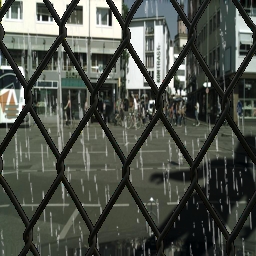}
        \end{minipage}%
        \begin{minipage}[c]{0.15\linewidth}
            \includegraphics[width=0.96\linewidth, height=0.48\linewidth]{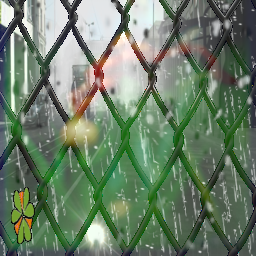}
        \end{minipage}%

        \begin{minipage}[c]{0.1\linewidth}
            \hfill 
            \text{\scriptsize BIDeN\;}
        \end{minipage}%
        \begin{minipage}[c]{0.15\linewidth}
            \includegraphics[width=0.96\linewidth, height=0.48\linewidth]{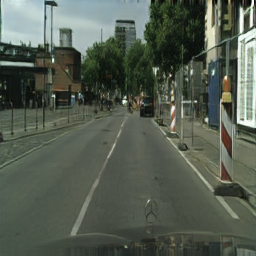}
        \end{minipage}%
        \begin{minipage}[c]{0.15\linewidth}
            \includegraphics[width=0.96\linewidth, height=0.48\linewidth]{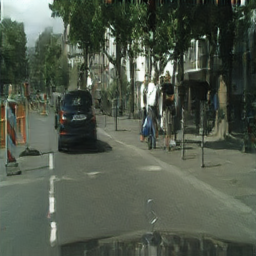}
        \end{minipage}%
        \begin{minipage}[c]{0.15\linewidth}
            \includegraphics[width=0.96\linewidth, height=0.48\linewidth]{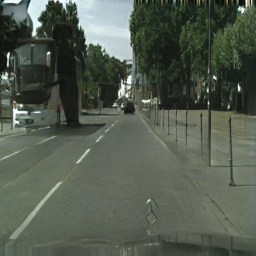}
        \end{minipage}%
        \begin{minipage}[c]{0.15\linewidth}
            \includegraphics[width=0.96\linewidth, height=0.48\linewidth]{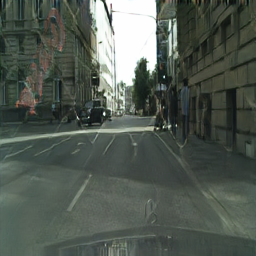}
        \end{minipage}%
        \begin{minipage}[c]{0.15\linewidth}
            \includegraphics[width=0.96\linewidth, height=0.48\linewidth]{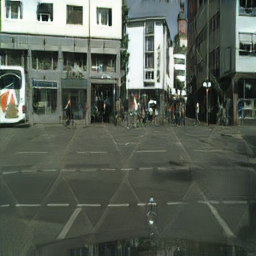}
        \end{minipage}%
        \begin{minipage}[c]{0.15\linewidth}
            \includegraphics[width=0.96\linewidth, height=0.48\linewidth]{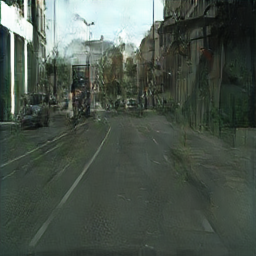}
        \end{minipage}%

        \begin{minipage}[c]{0.1\linewidth}
            \hfill 
            \text{\scriptsize CBDNet\;}
        \end{minipage}%
        \begin{minipage}[c]{0.15\linewidth}
            \includegraphics[width=0.96\linewidth, height=0.48\linewidth]{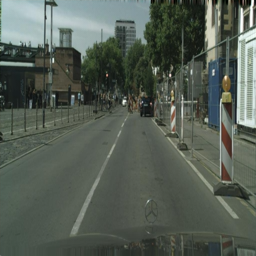}
        \end{minipage}%
        \begin{minipage}[c]{0.15\linewidth}
            \includegraphics[width=0.96\linewidth, height=0.48\linewidth]{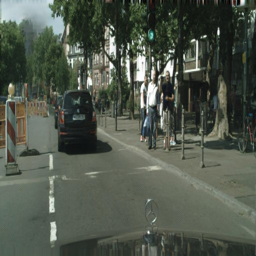}
        \end{minipage}%
        \begin{minipage}[c]{0.15\linewidth}
            \includegraphics[width=0.96\linewidth, height=0.48\linewidth]{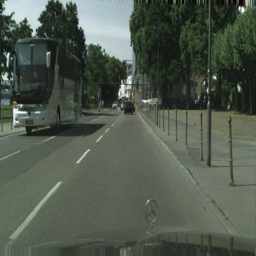}
        \end{minipage}%
        \begin{minipage}[c]{0.15\linewidth}
            \includegraphics[width=0.96\linewidth, height=0.48\linewidth]{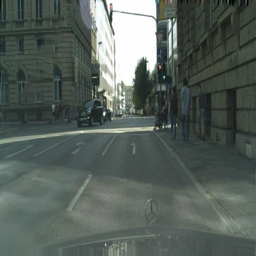}
        \end{minipage}%
        \begin{minipage}[c]{0.15\linewidth}
            \includegraphics[width=0.96\linewidth, height=0.48\linewidth]{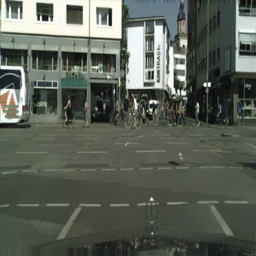}
        \end{minipage}%
        \begin{minipage}[c]{0.15\linewidth}
            \includegraphics[width=0.96\linewidth, height=0.48\linewidth]{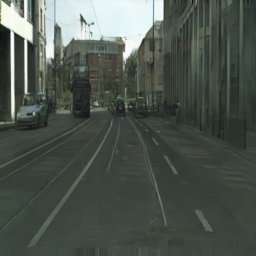}
        \end{minipage}%

        \begin{minipage}[c]{0.1\linewidth}
            \hfill 
            \text{\scriptsize GT\;}
        \end{minipage}%
        \begin{minipage}[c]{0.15\linewidth}
            \includegraphics[width=0.96\linewidth, height=0.48\linewidth]{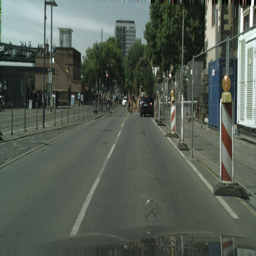}
        \end{minipage}%
        \begin{minipage}[c]{0.15\linewidth}
            \includegraphics[width=0.96\linewidth, height=0.48\linewidth]{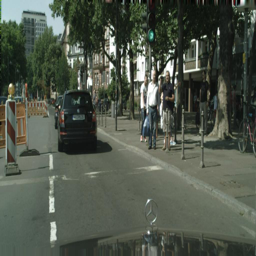}
        \end{minipage}%
        \begin{minipage}[c]{0.15\linewidth}
            \includegraphics[width=0.96\linewidth, height=0.48\linewidth]{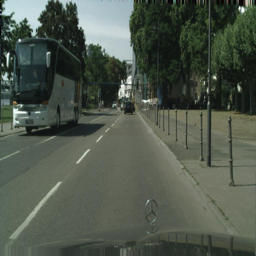}
        \end{minipage}%
        \begin{minipage}[c]{0.15\linewidth}
            \includegraphics[width=0.96\linewidth, height=0.48\linewidth]{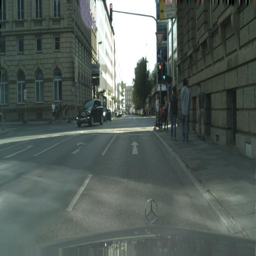}
        \end{minipage}%
        \begin{minipage}[c]{0.15\linewidth}
            \includegraphics[width=0.96\linewidth, height=0.48\linewidth]{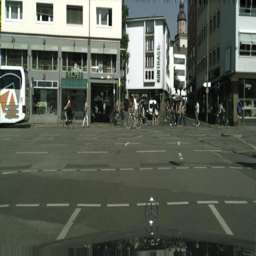}
        \end{minipage}%
        \begin{minipage}[c]{0.15\linewidth}
            \includegraphics[width=0.96\linewidth, height=0.48\linewidth]{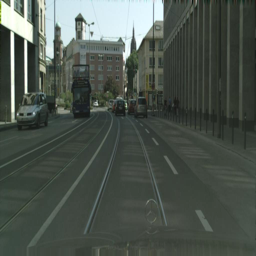}
        \end{minipage}%

        \begin{minipage}[c]{0.1\linewidth}
            \hfill 
            \text{\scriptsize \;}
        \end{minipage}%
        \begin{minipage}[c]{0.15\linewidth}
            \centering 
            \text{\scriptsize Case (1)}
        \end{minipage}%
        \begin{minipage}[c]{0.15\linewidth}
            \centering 
            \text{\scriptsize Case (2)}
        \end{minipage}%
        \begin{minipage}[c]{0.15\linewidth}
            \centering 
            \text{\scriptsize Case (3)}
        \end{minipage}%
        \begin{minipage}[c]{0.15\linewidth}
            \centering 
            \text{\scriptsize Case (4)}
        \end{minipage}%
        \begin{minipage}[c]{0.15\linewidth}
            \centering 
            \text{\scriptsize Case (5)}
        \end{minipage}%
        \begin{minipage}[c]{0.15\linewidth}
            \centering 
            \text{\scriptsize Case (6)}
        \end{minipage}%

    \end{minipage}
    \caption{\textbf{Qualitative results of Task III.A : Image restoration in multi-degradation removal.} Details of all cases are shown in Table~\ref{tab:task3-a}. CBDNet surpasses BIDeN in all cases, effectively restoring images with low visibility.}
    \label{fig:task3-a-fig}

\end{figure*}
\begin{table}[!thp]
  \centering
  \setlength{\tabcolsep}{1pt}
  \fontsize{7}{3}\selectfont
       \caption{\textbf{Quantitative results on Task III.A: Image restoration in multi-degradation removal.} We compare BIDeN and our method under 6 cases: (1) rain streak (2) rain streak + snow + moderate haze + raindrop (3) flare+reflection+shadow (4) fence + watermark (5) rain streak + shadow + fence (6) rain streak + snow + moderate haze + raindrop + flare + shadow + fence + reflection + watermark.}
    \begin{tabular}{lcc|cc|cc|cc|cc|cc}
    \toprule
     \Rows{Method}& \multicolumn{2}{c|}{Case (1)} & \multicolumn{2}{c|}{Case (2)} & \multicolumn{2}{c}{Case (3)} & \multicolumn{2}{c|}{Case (4)} & \multicolumn{2}{c|}{Case (5)} & \multicolumn{2}{c}{Case (6)}\cr
    &PSNR$\uparrow$ & SSIM$\uparrow$ &PSNR$\uparrow$ & SSIM$\uparrow$ &PSNR$\uparrow$ & SSIM$\uparrow$ &PSNR$\uparrow$ & SSIM$\uparrow$ &PSNR$\uparrow$ & SSIM$\uparrow$ &PSNR$\uparrow$ & SSIM$\uparrow$ \cr
    \midrule
    \multicolumn{1}{c}{Input}& 25.69 & 0.751 & 12.36 & 0.413  & 16.81 &0.789 & 18.13 & 0.640 & 17.21 & 0.487  & 10.42 &0.241\cr
    \multicolumn{1}{c}{BIDeN}& 27.36 & 0.858 & 24.12 & 0.778  & 24.45 &0.826 & 25.62 & 0.821 & 25.00 & 0.799  & 20.79 &0.658\cr
    \multicolumn{1}{c}{Ours}& \textbf{39.50} & \textbf{0.985} & \textbf{29.89} & \textbf{0.925}  & \textbf{32.44} &\textbf{0.964} & \textbf{32.12} & \textbf{0.956} & \textbf{31.23} & \textbf{0.944}  & \textbf{25.36} &\textbf{0.839}\cr
	\bottomrule
	\end{tabular}
	\label{tab:task3-a}
\end{table}

\noindent\textbf{Task II: Real-world bad weather removal.}
To validate the generalizability of our proposed method in real-world scenarios, we conduct experiments on natural images featuring rain streak, raindrop, and snow. 
All methods are trained on the synthetic dataset incorporating masks from rain streak, raindrop, and snow. We report the quantitative results with no-reference metrics, NIQE \cite{Mittal_Soundararajan_Bovik_2013} and BRISQUE \cite{Mittal_Moorthy_Bovik_2012} in Table~\ref{tab:task2}. BIDeN \cite{han2022bid} overall outperforms MPRNet \cite{Zamir_Arora_Khan_Hayat_Khan_Yang_Shao_2021}, particularly in the scenarios involving raindrop and snow. Benefiting from extensive self-supervised pre-training, CPNet \cite{Wang_2023_CVPR} exhibits exceptional generalization performance in this task. Despite CBDNet being trained from scratch, it still demonstrates performance comparable to CPNet. This suggests the substantial generalizability and robustness of our proposed method. In Figure~\ref{fig:task2-fig}, we present the qualitative results of real-world bad weather removal. In cases (1) to (6), the BIDeN method partially removed the weather elements, yet significant residual weather components were still evident. In contrast, CBDNet effectively eliminated the weather components. However, in case (5), CBDNet erroneously removed portions of the tree branches. In case (7), BIDeN demonstrated superior performance in removing snow. For case (8), both BIDeN and CBDNet showed inadequate results in processing rain streaks with splashing effects.

\subsection{Effectiveness of CBDNet for multi-domain and controllable BID}
We use the newly constructed multi-degradation removal dataset and perform controllable BID experiments. 

\begin{figure}[t]
\includegraphics[width=\linewidth]{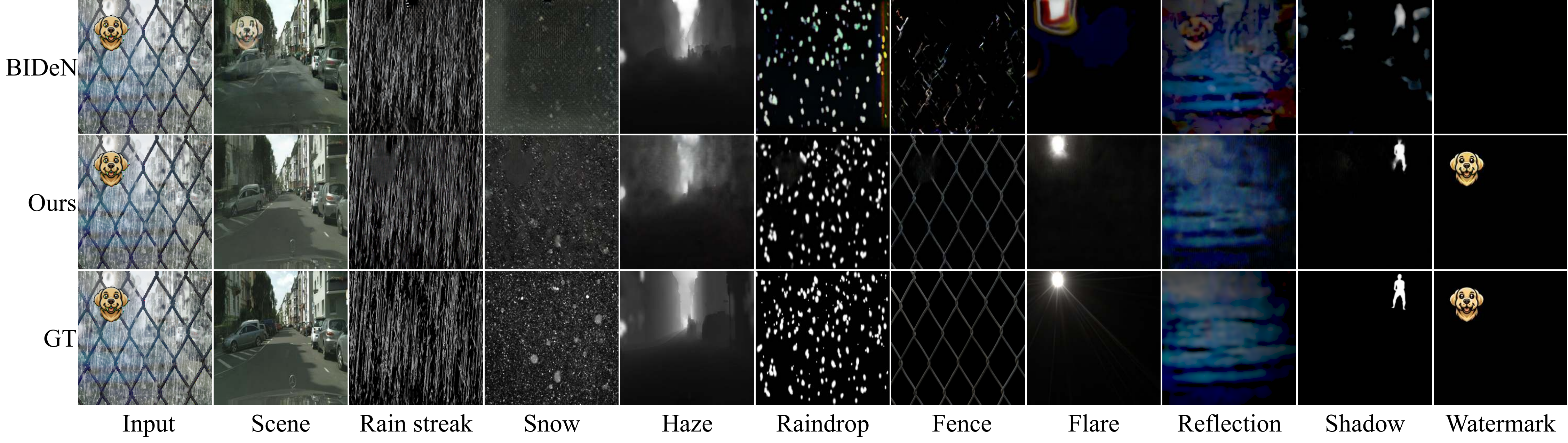}
\caption{\textbf{Qualitative results of Task III.B: Degradation masks reconstruction in multi-degradation removel.} The input images contains all nine degradations. BIDeN is not able to effectively separate the various degradation components, which is reflected in the failed reconstruction of the degradation masks, such as fence, shadow, and watermark. This also affects the outcome of image restoration, with residues of fence, watermarks, and shadows remaining on the scene image. CBDNet, on the other hand, successfully separates and reconstructs all components.}
\label{fig:task3-b-1}
\end{figure}
\begin{table}[!htb]
\centering
\fontsize{7}{3}\selectfont

\caption{\textbf{Quantitative results of Task III.B: Degradation masks reconstruction in multi-degradation removal.} The inputs contain all nine degradations.}
\label{tab:task3-b}
\begin{tabular}{lc|c|c|c|c|c|c|c|c|c}
\toprule
\multicolumn{1}{c}{Method} &\multicolumn{1}{c|}{Metric} & Rain Streak & Snow & Haze & Raindrop & Fence &Flare &Reflection&Shadow&Watermark\cr
\midrule
\raisebox{-1.0\height}{BIDeN} 
& PSNR$\uparrow$ & 24.95 & 20.52 & 23.75 & 17.70 &22.46&19.38&29.84&24.07&21.89 \cr
& SSIM$\uparrow$ & 0.651 & 0.292 & 0.861 & 0.517&0.202&0.893&\textbf{0.966}&\textbf{0.788}&0.933\cr
\midrule
\raisebox{-1.0\height}{Ours} 
& PSNR$\uparrow$ & \textbf{26.70} & \textbf{23.75} & \textbf{26.71} & \textbf{20.96} & \textbf{31.70} & \textbf{27.13} & \textbf{32.51} & \textbf{25.70} & \textbf{39.72}\cr
& SSIM$\uparrow$ & \textbf{0.698} & \textbf{0.643} & \textbf{0.868} & \textbf{0.799}& \textbf{0.824} & \textbf{0.937} & 0.961 & 0.670 & \textbf{0.988}\cr

\bottomrule
\end{tabular}

\end{table}

\noindent\textbf{Task III: Multi-degradation removal.}
This task involves a wide variety of degradations, including weather-related factors, lighting, and occlusions.
Considering the complexity of the task, we present our results through three subsections: image restoration, mask reconstruction, and controllable blind image decomposition. Apart from BIDeN, the other methods involved in the comparison previously cannot reconstruct the degradation mask and therefore do not participate in the comparison for this task.

\noindent\textbf{A: Image restoration.}
We report results for 6 cases in Figure~\ref{fig:task3-a-fig} and Table~\ref{tab:task3-a}. CBDNet consistently demonstrates superior image restoration capability compared to BIDeN. Notably, in the highly challenging case 5, BIDeN's output exhibits significant distortion, while CBDNet's results retain high quality, showing the strong capacity of CBDNet in multi-degradation removal.

\noindent\textbf{B: Degradation Masks Reconstruction.} We present mask reconstruction results for rain-streak, snow, haze, and raindrop under the case with all nine degradations in Table~\ref{tab:task3-b} and Figure~\ref{fig:task3-b-1}.  CBDNet outperforms BIDeN in most scenarios, showcasing its superior ability to remove multiple degradations.

\begin{figure*}[ht]
    \begin{minipage}{1\textwidth}
    \centering
        \begin{minipage}[c]{0.06\linewidth}
            \hfill 
            \text{\scriptsize Input}
        \end{minipage}%
        \begin{minipage}[c]{0.133\linewidth}
            \includegraphics[width=0.98\linewidth, height=0.49\linewidth]{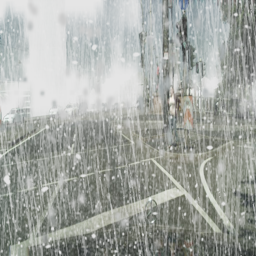}
        \end{minipage}%
        \begin{minipage}[c]{0.133\linewidth}
            \includegraphics[width=0.98\linewidth, height=0.49\linewidth]{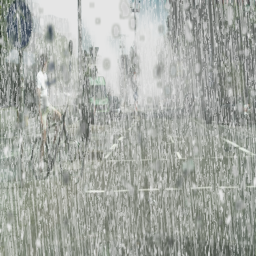}
        \end{minipage}%
        \begin{minipage}[c]{0.133\linewidth}
            \includegraphics[width=0.98\linewidth, height=0.49\linewidth]{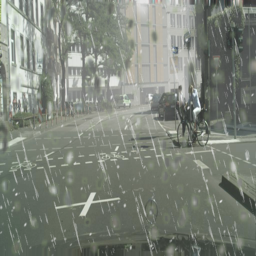}
        \end{minipage}%
        \begin{minipage}[c]{0.133\linewidth}
            \includegraphics[width=0.98\linewidth, height=0.49\linewidth]{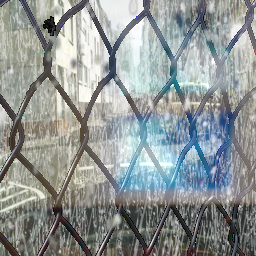}
        \end{minipage}%
        \begin{minipage}[c]{0.133\linewidth}
            \includegraphics[width=0.98\linewidth, height=0.49\linewidth]{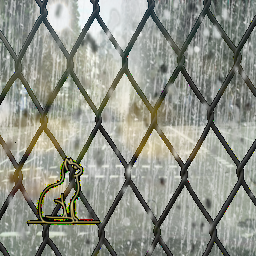}
        \end{minipage}%
        \begin{minipage}[c]{0.133\linewidth}
            \includegraphics[width=0.98\linewidth, height=0.49\linewidth]{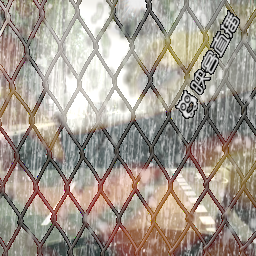}
        \end{minipage}%
        \begin{minipage}[c]{0.133\linewidth}
            \includegraphics[width=0.98\linewidth, height=0.49\linewidth]{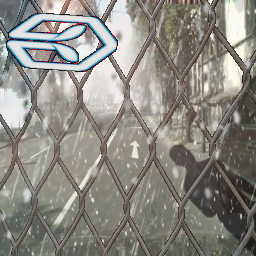}
        \end{minipage}%
        
        \begin{minipage}[c]{0.06\linewidth}
            \hfill 
            \text{\scriptsize Ours}
        \end{minipage}%
        \begin{minipage}[c]{0.133\linewidth}
            \includegraphics[width=0.98\linewidth, height=0.49\linewidth]{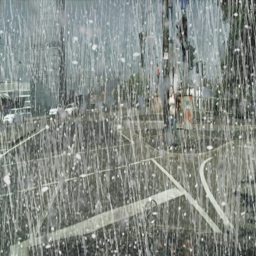}
        \end{minipage}%
        \begin{minipage}[c]{0.133\linewidth}
            \includegraphics[width=0.98\linewidth, height=0.49\linewidth]{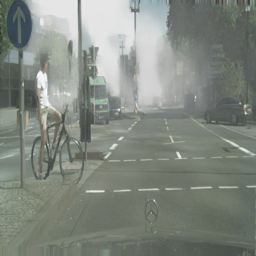}
        \end{minipage}%
        \begin{minipage}[c]{0.133\linewidth}
            \includegraphics[width=0.98\linewidth, height=0.49\linewidth]{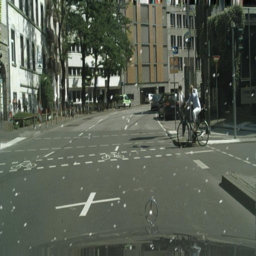}
        \end{minipage}%
        \begin{minipage}[c]{0.133\linewidth}
            \includegraphics[width=0.98\linewidth, height=0.49\linewidth]{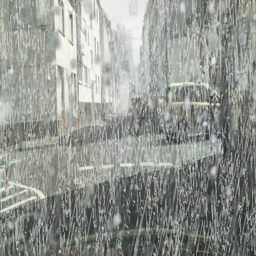}
        \end{minipage}%
        \begin{minipage}[c]{0.133\linewidth}
            \includegraphics[width=0.98\linewidth, height=0.49\linewidth]{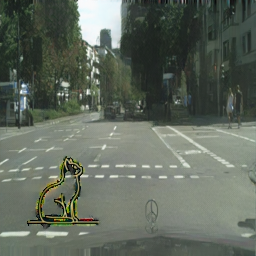}
        \end{minipage}%
        \begin{minipage}[c]{0.133\linewidth}
            \includegraphics[width=0.98\linewidth, height=0.49\linewidth]{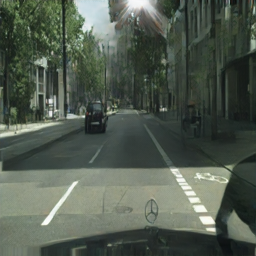}
        \end{minipage}%
        \begin{minipage}[c]{0.133\linewidth}
            \includegraphics[width=0.98\linewidth, height=0.49\linewidth]{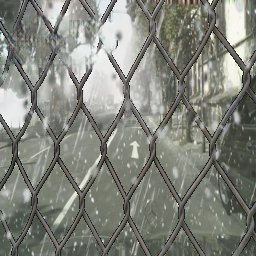}
        \end{minipage}%

        \begin{minipage}[c]{0.06\linewidth}
            \hfill 
            \text{\scriptsize GT}
        \end{minipage}%
        \begin{minipage}[c]{0.133\linewidth}
            \includegraphics[width=0.98\linewidth, height=0.49\linewidth]{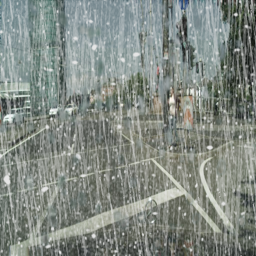}
        \end{minipage}%
        \begin{minipage}[c]{0.133\linewidth}
            \includegraphics[width=0.98\linewidth, height=0.49\linewidth]{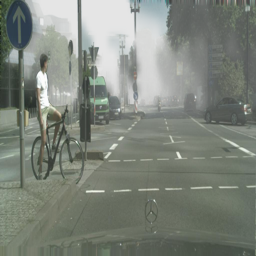}
        \end{minipage}%
        \begin{minipage}[c]{0.133\linewidth}
            \includegraphics[width=0.98\linewidth, height=0.49\linewidth]{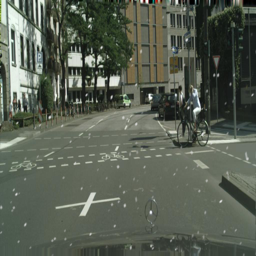}
        \end{minipage}%
        \begin{minipage}[c]{0.133\linewidth}
            \includegraphics[width=0.98\linewidth, height=0.49\linewidth]{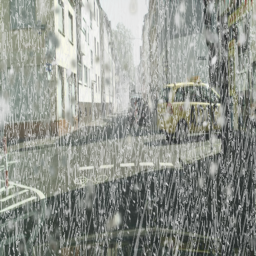}
        \end{minipage}%
        \begin{minipage}[c]{0.133\linewidth}
            \includegraphics[width=0.98\linewidth, height=0.49\linewidth]{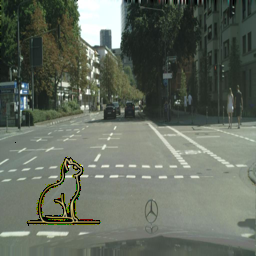}
        \end{minipage}%
        \begin{minipage}[c]{0.133\linewidth}
            \includegraphics[width=0.98\linewidth, height=0.49\linewidth]{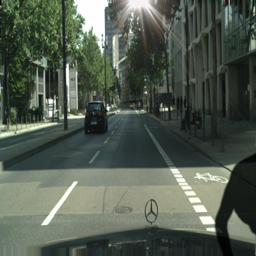}
        \end{minipage}%
        \begin{minipage}[c]{0.133\linewidth}
            \includegraphics[width=0.98\linewidth, height=0.49\linewidth]{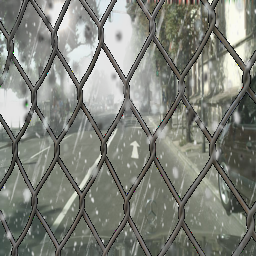}
        \end{minipage}%

        \begin{minipage}[c]{0.06\linewidth}
            \hfill 
            \text{\scriptsize \;}
        \end{minipage}%
        \begin{minipage}[c]{0.133\linewidth}
            \centering 
            \text{\scriptsize Prompt (1)}
        \end{minipage}%
        \begin{minipage}[c]{0.133\linewidth}
            \centering 
            \text{\scriptsize Prompt (2)}
        \end{minipage}%
        \begin{minipage}[c]{0.133\linewidth}
            \centering 
            \text{\scriptsize Prompt (3)}
        \end{minipage}%
        \begin{minipage}[c]{0.133\linewidth}
            \centering 
            \text{\scriptsize Prompt (4)}
        \end{minipage}%
        \begin{minipage}[c]{0.133\linewidth}
            \centering 
            \text{\scriptsize Prompt (5)}
        \end{minipage}%
        \begin{minipage}[c]{0.133\linewidth}
            \centering 
            \text{\scriptsize Prompt (6)}
        \end{minipage}%
        \begin{minipage}[c]{0.133\linewidth}
            \centering 
            \text{\scriptsize Prompt (7)}
        \end{minipage}%
    \end{minipage}

    \caption{\textbf{Qualitative results of Task III.C: Controllable blind image decomposition in multi-degradation removal.} Degradations of raindrop, haze, snow, and rain streak are presented in the input for Prompts (1)-(3), and all the nine degradations are present in the inputs for Prompts (4)-(6). Prompt (1): remove the haze and the raindrop; Prompt (2): generate the foggy cityscape; Prompt (3): retain the scene image and the snow; Prompt (4): keep the background and all weather components; Prompt (5): compose image with watermark; Prompt (6): restore scene, flare and shadow together; Prompt (7): remove watermark, reflection and shadow. Our method effectively removes the specified degradations as per the prompts, and the output images closely resemble the ideal post-processed versions.}
    \label{fig:task3-c-fig}
\end{figure*}

\begin{table}[!htb]
  \centering
    \fontsize{7}{3}\selectfont

  \caption{\textbf{Quantitative results of Task III.C: Controllable blind image decomposition in multi-degradation removal.} We use PSNR and SSIM metrics. For the meaning of Prompt (1)-(7), please refer to Fig~\ref{fig:task3-c-fig}.  This indicates that our approach learns strong representations for controllability BID.}
  \begin{tabular}{l|c|c|c|c|c|c|c}
    \toprule
    \multicolumn{1}{c|}{Metric} & Prompt (1) & Prompt (2) & Prompt (3) & Prompt (4) & Prompt (5) & Prompt (6) & Prompt (7)\\
    \midrule
        PSNR$\uparrow$ & 33.65 & 30.28 & 29.46 & 26.82 & 25.26 & 25.48 &28.40\\
        SSIM$\uparrow$ & 0.949 & 0.965 & 0.920 & 0.866 & 0.845 & 0.842 &0.961\\
    \bottomrule
  \end{tabular}

  \label{tab:task3-c}
\end{table}

\noindent\textbf{C: Controllable blind image decomposition.}
We demonstrate our method's capability to address specific types of degradation with six prompts. Results are shown in Table~\ref{tab:task3-c} and Figure~\ref{fig:task3-c-fig}. CBDNet adeptly removes specified degradation components and preserves the remaining ones as per the prompts.

\begin{figure}[htb!]
\includegraphics[width=\linewidth]{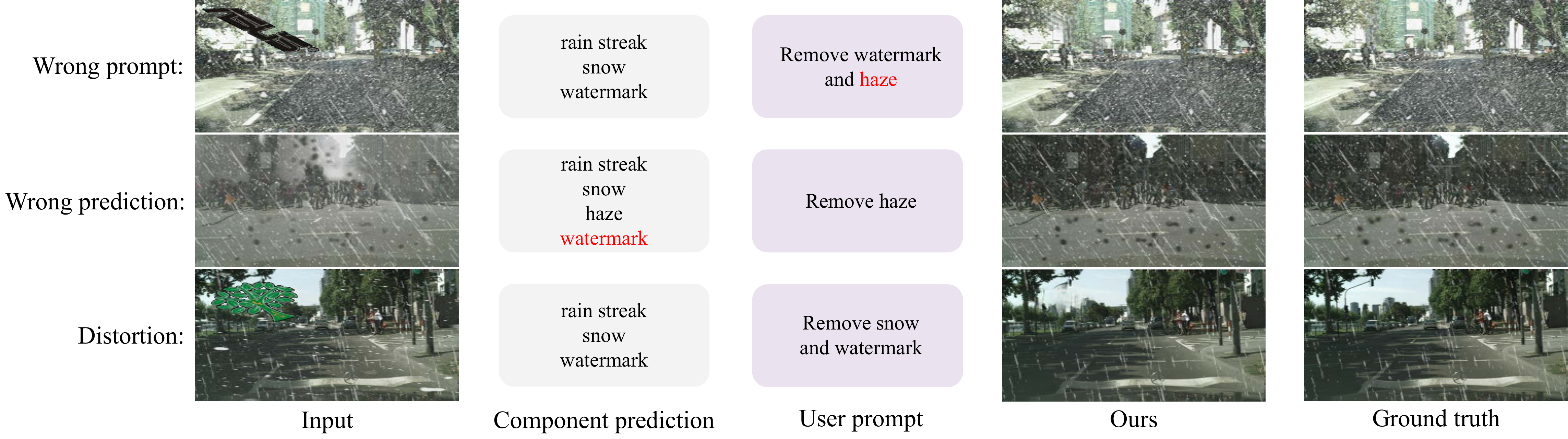}
\caption{\textbf{Qualitative results of corner cases.} Row 1 showcases a scenario in which the user's prompt includes an error: the instruction to remove haze is invalid since the input image does not contain any haze. Despite this, our method was not misled by the erroneous instruction and accurately executed watermark removal. Row 2 presents a case where the source classifier misidentified the image's components. Nonetheless, our method adhered to the user's directive to remove haze correctly. Row 3 reveals the limitations of CBDNet's restoration ability in heavily obscured areas. While our method effectively removed snow and watermark, it left noticeable distortions in the area previously occupied by the watermark.}
\label{fig:edge}
\end{figure}

We further visualize some corner-case results in Figure~\ref{fig:edge}. In the first case, CBDNet adeptly preserves rain streaks and snow while successfully removing the watermark. Notably, it maintains image quality by avoiding the erroneous retention of non-existent haze. In the second example, the source classifier incorrectly predicts that the image contains watermark components. Nonetheless, CBDNet can still accurately execute the user's instruction. This is because source prediction serves as merely auxiliary information. It allows users to more clearly understand the degradations contained in the image and does not impact the model's strict adherence to the user's prompts for completing specific image processing tasks. Moreover, our source classifier is highly effective, achieving over 99\% accuracy in the multi-weather removal task and 97\% in the multi-degradation removal task. Therefore, the probability of making a wrong prediction is very low. The third example illustrates that while CBDNet effectively removes obstructions, its in-painting capability has certain limitations, particularly when filling in extensively obscured areas.

\subsection{Ablation and Variant Study}

In Table~\ref{tab:ab}, we present the results of our ablation studies on cases (1) and (6). Comparing our baseline Restormer* with Model I (hyperparameter optimizations), it is evident that the hyperparameter optimizations applied to Restormer* significantly enhance its performance on Blind Image Decomposition (BID) tasks. Model II (loss selections) demonstrates the effectiveness of the chosen loss combination. Model III (decomposition block) shows that the decomposition block not only improves image restoration capabilities but also empowers the model with the ability to separate and reconstruct specific components. Furthermore, CBDNet (full model) with the recombination block and controllability block enables the model to perform controllable blind image decomposition under user prompts.

Model II represents a robust baseline for image restoration, optimized through diverse engineering techniques. In comparison with Model II, CBDNet's innovative decomposition and recombination blocks facilitate blind image decomposition without increasing parameter count, thereby enabling the model to adaptively decompose images as per user prompts. Moreover, the improvement of CBDNet on multi-degradation image restoration tasks is statistically highly significant ($p < 0.001$), derived from five repeated comparative experiments.

\begin{table}[!htb]
\centering
\fontsize{7}{3}\selectfont
\setlength{\tabcolsep}{6pt}
\caption{Quantitative results of ablation study on case (3) and (6). Details of all cases are shown in Table~\ref{tab:task1-a}. Restormer* represents our re-implementation of Restormer \cite{Zamir_Arora_Khan_Hayat_Khan_Yang_2022} without using progressive learning. Model I improves Restormer* with hyper-parameters optimizations. Model II uses Smooth L1 loss, VGG loss and LPIPS loss. Model III incorporates our proposed decomposition block, thereby gaining the capability for blind image decomposition. The final CBDNet builds on Model III with the addition of the proposed recombination block and the controllability block, allowing for the selective removal of specified degradations based on prompts.}

\begin{tabular}{lcc|cc|c|c}
\toprule
\multicolumn{1}{c|}{Ablation} & \multicolumn{2}{c|}{Case (3)} & \multicolumn{2}{c|}{Case (6)} & {\space BID \space} & Controllable BID \cr
& PSNR$\uparrow$ & SSIM$\uparrow$ & PSNR$\uparrow$ & SSIM$\uparrow$ & & \cr

\midrule
\multicolumn{1}{c|}{Restormer*} &32.15 &0.954 &27.19 &0.863 &- &-\cr
\multicolumn{1}{c|}{I} &33.78 &0.968 &29.19 &0.903 &- &-\cr
\multicolumn{1}{c|}{II}  &35.44 &0.983 &31.21 &0.949 &- &-\cr
\multicolumn{1}{c|}{III}  &35.82 &0.984 &31.52 &0.950 &\checkmark &-\cr
\midrule
\multicolumn{1}{c|}{CBDNet} &\textbf{36.00} &\textbf{0.985} &\textbf{31.55} &\textbf{0.952} &\checkmark &\checkmark\cr
\bottomrule
\end{tabular}
\vspace{1em}
\label{tab:ab}
\end{table}

\begin{table}[!htb]
  \centering
    \fontsize{7}{3}\selectfont
    \setlength{\tabcolsep}{8pt}
    \caption{\textbf{Quantitative results of substitution experiment.} The input images of this weather removal task contain rain streak, snow, moderate haze, and raindrop. For the decomposition block, we constructed variants of the decomposition block using both CNNs and transformers, which can similarly decompose feature maps into sub-feature maps corresponding to degradations. Similarly, we also built variants of the recombination block using CNNs and transformers, enabling effective blending of preserved components based on the sub-feature maps. All variant methods are capable of effectively accomplishing the controllable BID task, demonstrating that the decomposition-recombination strategy is effective. Considering efficiency, we opt for a parameter-free split and multiplication approach as part of CBDNet.}
  \begin{tabular}{l|c|c|cc|c}
    \toprule
    Method&Decomposition&Recombination&\multicolumn{2}{c|}{Weather removal}& Sum of params\cr
    & & & PSNR$\uparrow$ & SSIM$\uparrow$ & $\downarrow$\cr

    \midrule
    \multicolumn{1}{c|}{I} &Transformer&Multiplication&31.33 &0.950 &0.047\cr
    \multicolumn{1}{c|}{II} &CNN&Multiplication&31.55 &0.950 &0.035\cr
   \multicolumn{1}{c|}{III} &Split&Transformer&31.43 &0.949&0.047\cr
    \multicolumn{1}{c|}{IV} &Split&CNN&31.31 &0.948 &0.115\cr
    \midrule
    \multicolumn{1}{c|}{CBDNet} &Split&Multiplication&\textbf{31.55} &\textbf{0.952} &\textbf{0}\cr
    \bottomrule
  \end{tabular}

  \label{tab:vr}
\end{table}

Additionally, we investigate the impact of varied design choices on the performance of the decomposition and recombination blocks. Two versions of each block were developed—one based on transformers and the other on CNNs. The performance outcomes of these four variants are presented in Table~\ref{tab:vr}. The results indicate a comparable performance level across all methods. Taking efficiency into consideration, we opt to use the parameter-free version in CBDNet.

\section{Conclusion}

This paper introduces a controllable blind image decomposition (BID) approach to meet the complex and diverse user demands in real-world image processing. We present CBDNet, which features efficient decomposition, controllability, and recombination blocks specifically designed for controllable BID. To facilitate further research on multi-degradation removal, we have curated a challenging dataset that includes nine types of degradation from three domains. CBDNet demonstrates a strong baseline capability in standard BID, as validated through extensive experiments in a range of BID tasks. Furthermore, experiments on controllable BID demonstrate that CBDNet can effectively follow user prompts to achieve the removal of specific components, all while maintaining high efficiency. We hope that our study contributes to the advancement of low-level vision systems in the era of foundation models.

\clearpage

\bibliographystyle{splncs04}
\bibliography{bibtex}

\begin{thebibliography}{10}
\providecommand{\url}[1]{\texttt{#1}}
\providecommand{\urlprefix}{URL }
\providecommand{\doi}[1]{https://doi.org/#1}

\bibitem{avrahami2022blended}
Avrahami, O., Lischinski, D., Fried, O.: Blended diffusion for text-driven editing of natural images. In: Proceedings of the IEEE/CVF Conference on Computer Vision and Pattern Recognition. pp. 18208--18218 (2022)

\bibitem{bai2023textir}
Bai, Y., Wang, C., Xie, S., Dong, C., Yuan, C., Wang, Z.: Textir: A simple framework for text-based editable image restoration (2023)

\bibitem{Berman_Treibitz_Avidan_2016}
Berman, D., Avidan, S., et~al.: Non-local image dehazing. In: Proceedings of the IEEE conference on computer vision and pattern recognition. pp. 1674--1682 (2016)

\bibitem{raindrop}
Blinn, J.F.: A generalization of algebraic surface drawing. ACM Trans. Graph.  \textbf{1}(3),  235–256 (jul 1982). \doi{10.1145/357306.357310}, \url{https://doi.org/10.1145/357306.357310}

\bibitem{Blinn_1982}
Blinn, J.F.: A generalization of algebraic surface drawing. ACM transactions on graphics (TOG)  \textbf{1}(3),  235--256 (1982)

\bibitem{brooks2008structural}
Brooks, A.C., Zhao, X., Pappas, T.N.: Structural similarity quality metrics in a coding context: exploring the space of realistic distortions. IEEE Transactions on image processing  \textbf{17}(8),  1261--1273 (2008)

\bibitem{brooks2022instructpix2pix}
Brooks, T., Holynski, A., Efros, A.A.: Instructpix2pix: Learning to follow image editing instructions. In: Proceedings of the IEEE/CVF Conference on Computer Vision and Pattern Recognition. pp. 18392--18402 (2023)

\bibitem{brooks2023instructpix2pix}
Brooks, T., Holynski, A., Efros, A.A.: Instructpix2pix: Learning to follow image editing instructions (2023)

\bibitem{Cai_Xu_Jia_Qing_Tao_2016}
Cai, B., Xu, X., Jia, K., Qing, C., Tao, D.: Dehazenet: An end-to-end system for single image haze removal. IEEE transactions on image processing  \textbf{25}(11),  5187--5198 (2016)

\bibitem{chen2022learning}
Chen, W.T., Huang, Z.K., Tsai, C.C., Yang, H.H., Ding, J.J., Kuo, S.Y.: Learning multiple adverse weather removal via two-stage knowledge learning and multi-contrastive regularization: Toward a unified model. In: Proceedings of the IEEE/CVF Conference on Computer Vision and Pattern Recognition. pp. 17653--17662 (2022)

\bibitem{9710813}
Chen, Z., Long, C., Zhang, L., Xiao, C.: Canet: A context-aware network for shadow removal. In: 2021 IEEE/CVF International Conference on Computer Vision (ICCV). pp. 4723--4732 (2021). \doi{10.1109/ICCV48922.2021.00470}

\bibitem{Cichocki_Amari_2002}
Cichocki, A., Amari, S.i.: Adaptive blind signal and image processing: learning algorithms and applications. John Wiley \& Sons (2002)

\bibitem{cordts2016cityscapes}
Cordts, M., Omran, M., Ramos, S., Rehfeld, T., Enzweiler, M., Benenson, R., Franke, U., Roth, S., Schiele, B.: The cityscapes dataset for semantic urban scene understanding. In: Proceedings of the IEEE conference on computer vision and pattern recognition. pp. 3213--3223 (2016)

\bibitem{Cun_Pun_Shi_2020}
Cun, X., Pun, C.M., Shi, C.: Towards ghost-free shadow removal via dual hierarchical aggregation network and shadow matting gan. In: Proceedings of the AAAI Conference on Artificial Intelligence. vol.~34, pp. 10680--10687 (2020)

\bibitem{devlin2019bert}
Devlin, J., Chang, M.W., Lee, K., Toutanova, K.: Bert: Pre-training of deep bidirectional transformers for language understanding (2019)

\bibitem{ding2019argan}
Ding, B., Long, C., Zhang, L., Xiao, C.: Argan: Attentive recurrent generative adversarial network for shadow detection and removal (2019)

\bibitem{Dong_Loy_He_Tang_2016}
Dong, C., Loy, C.C., He, K., Tang, X.: Image super-resolution using deep convolutional networks. IEEE transactions on pattern analysis and machine intelligence  \textbf{38}(2),  295--307 (2015)

\bibitem{Du_Xu_Qiu_Zhen_Zhang_2020}
Du, Y., Xu, J., Qiu, Q., Zhen, X., Zhang, L.: Variational image deraining. In: Proceedings of the IEEE/CVF Winter Conference on Applications of Computer Vision. pp. 2406--2415 (2020)

\bibitem{Fan_Chen_Yuan_Hua_Yu_Chen_2018}
Fan, Q., Chen, D., Yuan, L., Hua, G., Yu, N., Chen, B.: Decouple learning for parameterized image operators. In: Proceedings of the European Conference on Computer Vision (ECCV). pp. 442--458 (2018)

\bibitem{gafni2022make}
Gafni, O., Polyak, A., Ashual, O., Sheynin, S., Parikh, D., Taigman, Y.: Make-a-scene: Scene-based text-to-image generation with human priors. In: European Conference on Computer Vision. pp. 89--106. Springer (2022)

\bibitem{Kun_Gai_Zhenwei_Shi_Changshui_Zhang_2008}
Gai, K., Shi, Z., Zhang, C.: Blindly separating mixtures of multiple layers with spatial shifts. In: 2008 IEEE Conference on Computer Vision and Pattern Recognition. pp.~1--8. IEEE (2008)

\bibitem{Gu_Meng_Zuo_Zhang_2017}
Gu, S., Meng, D., Zuo, W., Zhang, L.: Joint convolutional analysis and synthesis sparse representation for single image layer separation. In: Proceedings of the IEEE International Conference on Computer Vision. pp. 1708--1716 (2017)

\bibitem{han2022bid}
Han, J., Li, W., Fang, P., Sun, C., Hong, J., Armin, M.A., Petersson, L., Li, H.: Blind image decomposition. In: European Conference on Computer Vision (ECCV) (2022)

\bibitem{5206515}
He, K., Sun, J., Tang, X.: Single image haze removal using dark channel prior. IEEE transactions on pattern analysis and machine intelligence  \textbf{33}(12),  2341--2353 (2010)

\bibitem{hertz2022prompt}
Hertz, A., Mokady, R., Tenenbaum, J., Aberman, K., Pritch, Y., Cohen-Or, D.: Prompt-to-prompt image editing with cross attention control. arXiv preprint arXiv:2208.01626  (2022)

\bibitem{huynh2008scope}
Huynh-Thu, Q., Ghanbari, M.: Scope of validity of psnr in image/video quality assessment. Electronics letters  \textbf{44}(13),  800--801 (2008)

\bibitem{Hyvärinen_Oja_2000}
Hyv{\"a}rinen, A., Oja, E.: Independent component analysis: algorithms and applications. Neural networks  \textbf{13}(4-5),  411--430 (2000)

\bibitem{9710914}
Jin, Y., Sharma, A., Tan, R.T.: Dc-shadownet: Single-image hard and soft shadow removal using unsupervised domain-classifier guided network. In: 2021 IEEE/CVF International Conference on Computer Vision (ICCV). pp. 5007--5016 (2021). \doi{10.1109/ICCV48922.2021.00498}

\bibitem{kawar2022imagic}
Kawar, B., Zada, S., Lang, O., Tov, O., Chang, H., Dekel, T., Mosseri, I., Irani, M.: Imagic: Text-based real image editing with diffusion models. In: Proceedings of the IEEE/CVF Conference on Computer Vision and Pattern Recognition. pp. 6007--6017 (2023)

\bibitem{kirillov2023segment}
Kirillov, A., Mintun, E., Ravi, N., Mao, H., Rolland, C., Gustafson, L., Xiao, T., Whitehead, S., Berg, A.C., Lo, W.Y., et~al.: Segment anything. arXiv preprint arXiv:2304.02643  (2023)

\bibitem{Kupyn_Martyniuk_Wu_Wang_2019}
Kupyn, O., Martyniuk, T., Wu, J., Wang, Z.: Deblurgan-v2: Deblurring (orders-of-magnitude) faster and better. In: Proceedings of the IEEE/CVF international conference on computer vision. pp. 8878--8887 (2019)

\bibitem{le2019shadow}
Le, H., Samaras, D.: Shadow removal via shadow image decomposition (2019)

\bibitem{Li_Peng_Wang_Xu_Feng_2017}
Li, B., Peng, X., Wang, Z., Xu, J., Feng, D.: Aod-net: All-in-one dehazing network. In: Proceedings of the IEEE international conference on computer vision. pp. 4770--4778 (2017)

\bibitem{Li_Liu_Hu_Wu_Lv_Peng}
Li, B., Liu, X., Hu, P., Wu, Z., Lv, J., Peng, X.: All-in-one image restoration for unknown corruption. In: Proceedings of the IEEE/CVF Conference on Computer Vision and Pattern Recognition. pp. 17452--17462 (2022)

\bibitem{Li_Cheong_Tan_2019}
Li, R., Cheong, L.F., Tan, R.T.: Heavy rain image restoration: Integrating physics model and conditional adversarial learning. In: Proceedings of the IEEE/CVF conference on computer vision and pattern recognition. pp. 1633--1642 (2019)

\bibitem{Li_Tan_Cheong_2020}
Li, R., Tan, R.T., Cheong, L.F.: All in one bad weather removal using architectural search. In: Proceedings of the IEEE/CVF conference on computer vision and pattern recognition. pp. 3175--3185 (2020)

\bibitem{Li_Araujo_Ren_Wang_Tokuda_Junior_Cesar-Junior_Zhang_Guo_Cao_2019}
Li, S., Araujo, I.B., Ren, W., Wang, Z., Tokuda, E.K., Junior, R.H., Cesar-Junior, R., Zhang, J., Guo, X., Cao, X.: Single image deraining: A comprehensive benchmark analysis. In: Proceedings of the IEEE/CVF Conference on Computer Vision and Pattern Recognition. pp. 3838--3847 (2019)

\bibitem{Li_Wu_Lin_Liu_Zha_2018}
Li, X., Wu, J., Lin, Z., Liu, H., Zha, H.: Recurrent squeeze-and-excitation context aggregation net for single image deraining. In: Proceedings of the European conference on computer vision (ECCV). pp. 254--269 (2018)

\bibitem{li2023promptinprompt}
Li, Z., Lei, Y., Ma, C., Zhang, J., Shan, H.: Prompt-in-prompt learning for universal image restoration (2023)

\bibitem{Liang_Cao_Sun_Zhang_Van_Gool_Timofte_2021}
Liang, J., Cao, J., Sun, G., Zhang, K., Van~Gool, L., Timofte, R.: Swinir: Image restoration using swin transformer. In: Proceedings of the IEEE/CVF international conference on computer vision. pp. 1833--1844 (2021)

\bibitem{Liu_Zhu_Bai_2021}
Liu, Y., Zhu, Z., Bai, X.: Wdnet: Watermark-decomposition network for visible watermark removal. In: Proceedings of the IEEE/CVF Winter Conference on Applications of Computer Vision. pp. 3685--3693 (2021)

\bibitem{Liu_Lai_Yang_Chuang_Huang_2020}
Liu, Y.L., Lai, W.S., Yang, M.H., Chuang, Y.Y., Huang, J.B.: Learning to see through obstructions with layered decomposition. IEEE transactions on pattern analysis and machine intelligence  \textbf{44}(11),  8387--8402 (2021)

\bibitem{Liu_Jaw_Huang_Hwang_2018}
Liu, Y.F., Jaw, D.W., Huang, S.C., Hwang, J.N.: Desnownet: Context-aware deep network for snow removal. IEEE Transactions on Image Processing  \textbf{27}(6),  3064--3073 (2018)

\bibitem{ma2023prores}
Ma, J., Cheng, T., Wang, G., Zhang, Q., Wang, X., Zhang, L.: Prores: Exploring degradation-aware visual prompt for universal image restoration (2023)

\bibitem{Mehri_Ardakani_Sappa_2021}
Mehri, A., Ardakani, P.B., Sappa, A.D.: Mprnet: Multi-path residual network for lightweight image super resolution. In: Proceedings of the IEEE/CVF Winter Conference on Applications of Computer Vision. pp. 2704--2713 (2021)

\bibitem{Mittal_Moorthy_Bovik_2012}
Mittal, A., Moorthy, A.K., Bovik, A.C.: No-reference image quality assessment in the spatial domain. IEEE Transactions on image processing  \textbf{21}(12),  4695--4708 (2012)

\bibitem{mittal2012no}
Mittal, A., Moorthy, A.K., Bovik, A.C.: No-reference image quality assessment in the spatial domain. IEEE Transactions on image processing  \textbf{21}(12),  4695--4708 (2012)

\bibitem{Mittal_Soundararajan_Bovik_2013}
Mittal, A., Soundararajan, R., Bovik, A.C.: Making a “completely blind” image quality analyzer. IEEE Signal processing letters  \textbf{20}(3),  209--212 (2012)

\bibitem{mittal2012making}
Mittal, A., Soundararajan, R., Bovik, A.C.: Making a “completely blind” image quality analyzer. IEEE Signal processing letters  \textbf{20}(3),  209--212 (2012)

\bibitem{Niu_Wen_Ren_Zhang_Yang_Wang_Zhang_Cao_Shen_2020}
Niu, B., Wen, W., Ren, W., Zhang, X., Yang, L., Wang, S., Zhang, K., Cao, X., Shen, H.: Single image super-resolution via a holistic attention network. In: Computer Vision--ECCV 2020: 16th European Conference, Glasgow, UK, August 23--28, 2020, Proceedings, Part XII 16. pp. 191--207. Springer (2020)

\bibitem{Pan_Liu_Sun_Zhang_Liu_Ren_Li_Tang_Lu_Tai_2018}
Pan, J., Liu, S., Sun, D., Zhang, J., Liu, Y., Ren, J., Li, Z., Tang, J., Lu, H., Tai, Y.W., et~al.: Learning dual convolutional neural networks for low-level vision. In: Proceedings of the IEEE conference on computer vision and pattern recognition. pp. 3070--3079 (2018)

\bibitem{10204770}
Park, D., Lee, B.H., Chun, S.Y.: All-in-one image restoration for unknown degradations using adaptive discriminative filters for specific degradations. In: 2023 IEEE/CVF Conference on Computer Vision and Pattern Recognition (CVPR). pp. 5815--5824 (2023). \doi{10.1109/CVPR52729.2023.00563}

\bibitem{Porav_Bruls_Newman_2019}
Porav, H., Bruls, T., Newman, P.: I can see clearly now: Image restoration via de-raining. In: 2019 International Conference on Robotics and Automation (ICRA). pp. 7087--7093. IEEE (2019)

\bibitem{Potlapalli_Zamir_Khan_Khan_2023}
Potlapalli, V., Zamir, S.W., Khan, S., Khan, F.S.: Promptir: Prompting for all-in-one blind image restoration. arXiv preprint arXiv:2306.13090  (2023)

\bibitem{Qian_Tan_Yang_Su_Liu_2018}
Qian, R., Tan, R.T., Yang, W., Su, J., Liu, J.: Attentive generative adversarial network for raindrop removal from a single image. In: Proceedings of the IEEE conference on computer vision and pattern recognition. pp. 2482--2491 (2018)

\bibitem{Quan_Yu_Liang_Yang_2021}
Quan, R., Yu, X., Liang, Y., Yang, Y.: Removing raindrops and rain streaks in one go. In: Proceedings of the IEEE/CVF Conference on Computer Vision and Pattern Recognition. pp. 9147--9156 (2021)

\bibitem{Quan_Deng_Chen_Ji_2019}
Quan, Y., Deng, S., Chen, Y., Ji, H.: Deep learning for seeing through window with raindrops. In: Proceedings of the IEEE/CVF International Conference on Computer Vision. pp. 2463--2471 (2019)

\bibitem{ramesh2022hierarchical}
Ramesh, A., Dhariwal, P., Nichol, A., Chu, C., Chen, M.: Hierarchical text-conditional image generation with clip latents. arXiv preprint arXiv:2204.06125  \textbf{1}(2), ~3 (2022)

\bibitem{Ren_Ma_Zhang_Pan_Cao_Liu_Yang_2018}
Ren, W., Ma, L., Zhang, J., Pan, J., Cao, X., Liu, W., Yang, M.H.: Gated fusion network for single image dehazing. In: Proceedings of the IEEE conference on computer vision and pattern recognition. pp. 3253--3261 (2018)

\bibitem{Russakovsky_Deng_Su_Krause_Satheesh_Ma_Huang_Karpathy_Khosla_Bernstein_et}
Russakovsky, O., Deng, J., Su, H., Krause, J., Satheesh, S., Ma, S., Huang, Z., Karpathy, A., Khosla, A., Bernstein, M., et~al.: Imagenet large scale visual recognition challenge. International journal of computer vision  \textbf{115},  211--252 (2015)

\bibitem{Sakaridis_Dai_Van_Gool_2018}
Sakaridis, C., Dai, D., Van~Gool, L.: Semantic foggy scene understanding with synthetic data. International Journal of Computer Vision  \textbf{126},  973--992 (2018)

\bibitem{Simonyan_Zisserman_2015}
Simonyan, K., Zisserman, A.: Very deep convolutional networks for large-scale image recognition. arXiv preprint arXiv:1409.1556  (2014)

\bibitem{Jose_Valanarasu_Yasarla_Patel_2022}
Valanarasu, J.M.J., Yasarla, R., Patel, V.M.: Transweather: Transformer-based restoration of images degraded by adverse weather conditions. In: Proceedings of the IEEE/CVF Conference on Computer Vision and Pattern Recognition. pp. 2353--2363 (2022)

\bibitem{Wang_2023_CVPR}
Wang, C., Zheng, Z., Quan, R., Sun, Y., Yang, Y.: Context-aware pretraining for efficient blind image decomposition. In: Proceedings of the IEEE/CVF Conference on Computer Vision and Pattern Recognition (CVPR). pp. 18186--18195 (June 2023)

\bibitem{Wang_Li_Yang_2018}
Wang, J., Li, X., Yang, J.: Stacked conditional generative adversarial networks for jointly learning shadow detection and shadow removal. In: Proceedings of the IEEE conference on computer vision and pattern recognition. pp. 1788--1797 (2018)

\bibitem{Wang_Yu_Wu_Gu_Liu_Dong_Qiao_Loy_2019}
Wang, X., Yu, K., Wu, S., Gu, J., Liu, Y., Dong, C., Qiao, Y., Change~Loy, C.: Esrgan: Enhanced super-resolution generative adversarial networks. In: Proceedings of the European conference on computer vision (ECCV) workshops. pp.~0--0 (2018)

\bibitem{Wang_Yu_Zhang_2022}
Wang, Y., Yu, J., Zhang, J.: Zero-shot image restoration using denoising diffusion null-space model. arXiv preprint arXiv:2212.00490  (2022)

\bibitem{Wang_Cun_Bao_Zhou_Liu_Li_2022}
Wang, Z., Cun, X., Bao, J., Zhou, W., Liu, J., Li, H.: Uformer: A general u-shaped transformer for image restoration. In: Proceedings of the IEEE/CVF conference on computer vision and pattern recognition. pp. 17683--17693 (2022)

\bibitem{Wu_He_Xue_Garg_Chen_Veeraraghavan_Barron_2021}
Wu, Y., He, Q., Xue, T., Garg, R., Chen, J., Veeraraghavan, A., Barron, J.T.: How to train neural networks for flare removal. In: Proceedings of the IEEE/CVF International Conference on Computer Vision. pp. 2239--2247 (2021)

\bibitem{yang2023languagedriven}
Yang, H., Pan, L., Yang, Y., Liang, W.: Language-driven all-in-one adverse weather removal (2023)

\bibitem{Yang_Tan_Feng_Guo_Yan_Liu_2020}
Yang, W., Tan, R.T., Feng, J., Guo, Z., Yan, S., Liu, J.: Joint rain detection and removal from a single image with contextualized deep networks. IEEE transactions on pattern analysis and machine intelligence  \textbf{42}(6),  1377--1393 (2019)

\bibitem{Yang_Tan_Feng_Liu_Guo_Yan_2017}
Yang, W., Tan, R.T., Feng, J., Liu, J., Guo, Z., Yan, S.: Deep joint rain detection and removal from a single image. In: Proceedings of the IEEE conference on computer vision and pattern recognition. pp. 1357--1366 (2017)

\bibitem{Yasarla_Sindagi_Patel_2020}
Yasarla, R., Sindagi, V.A., Patel, V.M.: Syn2real transfer learning for image deraining using gaussian processes. In: Proceedings of the IEEE/CVF conference on computer vision and pattern recognition. pp. 2726--2736 (2020)

\bibitem{Zamir_Arora_Khan_Hayat_Khan_Yang_2022}
Zamir, S.W., Arora, A., Khan, S., Hayat, M., Khan, F.S., Yang, M.H.: Restormer: Efficient transformer for high-resolution image restoration. In: Proceedings of the IEEE/CVF conference on computer vision and pattern recognition. pp. 5728--5739 (2022)

\bibitem{Zamir_Arora_Khan_Hayat_Khan_Yang_Shao_2021}
Zamir, S.W., Arora, A., Khan, S., Hayat, M., Khan, F.S., Yang, M.H., Shao, L.: Multi-stage progressive image restoration. In: Proceedings of the IEEE/CVF conference on computer vision and pattern recognition. pp. 14821--14831 (2021)

\bibitem{Zhang_Patel_2018}
Zhang, H., Patel, V.M.: Densely connected pyramid dehazing network. In: Proceedings of the IEEE conference on computer vision and pattern recognition. pp. 3194--3203 (2018)

\bibitem{Zhang_Sindagi_Patel_2020}
Zhang, H., Sindagi, V., Patel, V.M.: Image de-raining using a conditional generative adversarial network. IEEE transactions on circuits and systems for video technology  \textbf{30}(11),  3943--3956 (2019)

\bibitem{Zhang_Ren_Zhang_Zhang_Nie_Xue_Cao_2022}
Zhang, J., Ren, W., Zhang, S., Zhang, H., Nie, Y., Xue, Z., Cao, X.: Hierarchical density-aware dehazing network. IEEE Transactions on Cybernetics  \textbf{52}(10),  11187--11199 (2021)

\bibitem{Zhang_Huang_Yao_Yang_Yu_Zhou_Zhao}
Zhang, J., Huang, J., Yao, M., Yang, Z., Yu, H., Zhou, M., Zhao, F.: Ingredient-oriented multi-degradation learning for image restoration. In: Proceedings of the IEEE/CVF Conference on Computer Vision and Pattern Recognition. pp. 5825--5835 (2023)

\bibitem{10204072}
Zhang, J., Huang, J., Yao, M., Yang, Z., Yu, H., Zhou, M., Zhao, F.: Ingredient-oriented multi-degradation learning for image restoration. In: 2023 IEEE/CVF Conference on Computer Vision and Pattern Recognition (CVPR). pp. 5825--5835 (2023). \doi{10.1109/CVPR52729.2023.00564}

\bibitem{Zhang_Li_Yu_Luo_Li_2021}
Zhang, K., Li, R., Yu, Y., Luo, W., Li, C.: Deep dense multi-scale network for snow removal using semantic and depth priors. IEEE Transactions on Image Processing  \textbf{30},  7419--7431 (2021)

\bibitem{zhang2023adding}
Zhang, L., Rao, A., Agrawala, M.: Adding conditional control to text-to-image diffusion models. In: Proceedings of the IEEE/CVF International Conference on Computer Vision. pp. 3836--3847 (2023)

\bibitem{Zhang_Isola_Efros_Shechtman_Wang_2018}
Zhang, R., Isola, P., Efros, A.A., Shechtman, E., Wang, O.: The unreasonable effectiveness of deep features as a perceptual metric. In: Proceedings of the IEEE conference on computer vision and pattern recognition. pp. 586--595 (2018)

\bibitem{Zhang_Ng_Chen_2018}
Zhang, X., Ng, R., Chen, Q.: Single image reflection separation with perceptual losses. In: Proceedings of the IEEE conference on computer vision and pattern recognition. pp. 4786--4794 (2018)

\bibitem{Zhang_Tian_Kong_Zhong_Fu_2018}
Zhang, Y., Tian, Y., Kong, Y., Zhong, B., Fu, Y.: Residual dense network for image super-resolution. In: Proceedings of the IEEE conference on computer vision and pattern recognition. pp. 2472--2481 (2018)

\bibitem{zhong2024languageguided}
Zhong, H., Hong, Y., Weng, S., Liang, J., Shi, B.: Language-guided image reflection separation (2024)

\bibitem{Zhu_Fu_Lischinski_Heng_2017}
Zhu, L., Fu, C.W., Lischinski, D., Heng, P.A.: Joint bi-layer optimization for single-image rain streak removal. In: Proceedings of the IEEE international conference on computer vision. pp. 2526--2534 (2017)

\bibitem{Zhu_Wang_Fu_Yang_Guo_Dai_Qiao_Hu}
Zhu, Y., Wang, T., Fu, X., Yang, X., Guo, X., Dai, J., Qiao, Y., Hu, X.: Learning weather-general and weather-specific features for image restoration under multiple adverse weather conditions. In: Proceedings of the IEEE/CVF Conference on Computer Vision and Pattern Recognition. pp. 21747--21758 (2023)

\bibitem{zou2023segment}
Zou, X., Yang, J., Zhang, H., Li, F., Li, L., Gao, J., Lee, Y.J.: Segment everything everywhere all at once. arXiv preprint arXiv:2304.06718  (2023)

\end{thebibliography}

\clearpage
\appendix
\section{Datasets}
\textbf{Task I: Blind Image Decomposition.} This dataset, constructed by Han \emph{et al.}~\cite{han2022bid}, serves as a benchmark for comparing the capabilities of methods for removing weather components in the BID setting. The real-world driving images are sourced from the CityScape \cite{cordts2016cityscapes} dataset. Masks are obtained from Foggy CityScape \cite{Sakaridis_Dai_Van_Gool_2018}, Rain100L/Rain100H datasets \cite{Yang_Tan_Feng_Liu_Guo_Yan_2017}, Snow100K dataset \cite{Liu_Jaw_Huang_Hwang_2018}. For raindrop masks, this dataset employs the metaball model to accurately represent the shape and characteristics of the droplets \cite{Blinn_1982}.

\noindent\textbf{Task II: Real-world bad weather removal.}
The natural images in the training set are selected from the datasets of \cite{Qian_Tan_Yang_Su_Liu_2018}, Snow100K \cite{Liu_Jaw_Huang_Hwang_2018}, and Rain1800 \cite{Yang_Tan_Feng_Liu_Guo_Yan_2017}.
The testing set comprises real-world images only, each featuring elements like rain streaks, raindrops, or snow, sourced respectively from \cite{Li_Araujo_Ren_Wang_Tokuda_Junior_Cesar-Junior_Zhang_Guo_Cao_2019}, \cite{Qian_Tan_Yang_Su_Liu_2018}, and \cite{Liu_Jaw_Huang_Hwang_2018}.

\noindent\textbf{Task III: Multi-degradation removal.} This dataset encompasses a wide range of degradations across various domains. In the weather domain, it includes rain streaks, raindrops, snow, and haze. The lighting domain features flare, reflections, and shadows, while the obstruction domain comprises fence and watermark degradations. The degradation masks for the weather domain are sourced from \cite{Yang_Tan_Feng_Liu_Guo_Yan_2017, Sakaridis_Dai_Van_Gool_2018, Liu_Jaw_Huang_Hwang_2018, Blinn_1982}, aligning with the dataset used in Task I. The shadow masks (9500 in the training set and 500 in the testing set) are obtained from ISTD \cite{Wang_Li_Yang_2018} and SRD \cite{Cun_Pun_Shi_2020}. The reflection synthesis algorithm follows that of \cite{Zhang_Ng_Chen_2018}, with 2580 training images and 540 testing images from its reflection dataset. The watermark masks (2460 in the training set and 540 in the testing set) are collected from LVM \cite{Liu_Zhu_Bai_2021}. The flare masks (2500 in the training set and 500 in the testing set) are adopted from \cite{Wu_He_Xue_Garg_Chen_Veeraraghavan_Barron_2021}. The fence masks (545 in the training set and 100 in the testing set) are collected from SOLD \cite{Liu_Lai_Yang_Chuang_Huang_2020}. The process of creating degraded images is based on physical imaging models, as proposed in \cite{Porav_Bruls_Newman_2019} and \cite{5206515}.

\section{Training details}
\textbf{CBDNet.} Our method is trained on one Nvidia RTX 4090 GPU for one day, and the CUDA version is 11.8. The Adam optimizer was used for optimization, with $\beta _1=0.5$ nad $\beta _2 = 0.999$. The training process consists of 80 epochs, where the learning rate starts at 0.0003 and gradually decreases according to a cosine annealing decay schedule. We initialized the weights of CBDNet using Xavier initialization and employed a batch size of 1. Both training and testing images were loaded at a resolution of $256 \times 256$. For data augmentation, we applied a random horizontal flip. To provide an equitable comparison, we use the same resolution and data augmentation in the following methods as CBDNet.\\
\textbf{Existing methods.}
We report results of Restormer \cite{Zamir_Arora_Khan_Hayat_Khan_Yang_2022} , MPRNet \cite{Mehri_Ardakani_Sappa_2021}, BIDeN \cite{han2022bid} and CPNet \cite{Wang_2023_CVPR} in our setting. Restormer \cite{Zamir_Arora_Khan_Hayat_Khan_Yang_2022} is trained on one Nvidia RTX4090 GPU for 80 epochs with a batch size of 1. MPRNet is trained on four Tesla P100-PCIE-16GB GPUs with 250 epochs and a batch size of 16. BIDeN \cite{han2022bid} is trained on one Tesla P100-PCIE-16GB GPU with 200 epochs and a batch size of 1. For CPNet \cite{Wang_2023_CVPR}, we report their official result. All these methods are trained with a learning rate of 0.0003, and other settings not mentioned above follow their official ones.

\section{Evaluation metrics}

We evaluate image quality using four general evaluation metrics. Peak Signal-to-Noise Ratio (PSNR) \cite{huynh2008scope} and Structural Similarity Index Measure (SSIM) \cite{brooks2008structural} are used for evaluating image quality with reference images. For non-reference images, the Blind/Referenceless Image Spatial Quality Evaluator (BRISQUE) \cite{mittal2012no} and Naturalness Image Quality Evaluator (NIQE) \cite{mittal2012making} are applied.

\section{Implementation details}
\label{sec:prompt}
\subsection{Image encoder}
Image encoder begins with a convolutional layer $embed\_conv$ (in features=3, out features=24, kernel size=3).  It is followed by four encoder blocks, consisting of multiple transformer blocks. These blocks progressively encode the input into a feature map $ F_d \in \mathcal{R}^{H\times W\times 40}$. The encoders are characterized by their parameters: num blocks=[4, 6, 6, 8], num heads=[1, 2, 4, 10], and channels=[24, 48, 48, 40]. Between each encoder block, down-sampling layers are deployed to ensure the model captures broader contextual information at each level.

\subsection{Decomposition block}
The entire operation of the decomposition module is a channel split operation, which evenly divides the feature map $ F_d \in \mathcal{R}^{H\times W\times 40}$ by the number of components $10$, resulting in the sub-feature maps list $ F_{c_{i}}\in \mathcal{R}^{H\times W\times 4} $.

\subsection{Controllability block}
\textbf{Source classifier} comprises several components: $layer_1$ (in features=20, out features=80, kernel size=3), $AvgPool_1$ (kernel size=2, stride=2), $layer_2$ (in features=80, out features=20, kernel size=3), $AvgPool_2$ (kernel size=2, stride=2), and $fc_1$ (in features=$\frac{H}{32}\times \frac{W}{32}\times 20$, out features=10). Batch normalization, ReLU activations are applied for stabilization and non-linearity.

\noindent\textbf{Prompt converter} employs a pre-trained BERT encoder to extract the embedding of a prompt. Following this, an attention classifier refines the relevant features for the classification task. Within the Attention Classifier, a linear layer $fc_1$ (in features=768, out features=64) first reduces the dimensionality of the text embedding. Subsequently, a MultiheadAttention module (num heads=16, dropout=0.1) captures complex dependencies and relationships within the embedding. Following this, the Attention Classifier utilizes two fully connected layers, $fc_2$ (in features=64, out features=32) and $fc_3$ (in features=32, out features=num classes), to categorize the embeddings into predefined classes. For the task of multi-degradation removal, the classifier distinguishes among 10 classes. To address the varying lengths of input sequences, an adaptive average pooling layer is applied after the attention classifier.

\subsection{Recombination block}
The core operation of the recombination module is multiplication and addition, multiplying the sub-feature maps with the categorical vector $V_c \in \mathcal{R}^{10}$ and then summing them up to obtain the mixed feature map $ F_r \in \mathcal{R}^{\frac{H}{8}\times \frac{W}{8}\times 4}$.

\subsection{Image decoder}
Mirroring the structure of the image encoder but in reverse, the decoders work to reconstruct the spatial dimensions of the input from the condensed feature maps through a series of transformer blocks. This reconstruction process is enhanced by up-sampling layers and reduction convolution layers, which merge features across different scales. Following the decoding phase, a refinement stage that includes 4 additional transformer blocks further processes the features. The final output is then generated through a sequence of convolutional layers: $layer_1$ (in features=48, out features=96, kernel size=3), $layer_2$ (in features=96, out features=32, kernel size=3), and back to $layer_1$ (in features=32, out features=3, kernel size=3). To ensure stability and introduce non-linearity, batch normalization and ReLU activations are employed throughout.

\section{Additional Results}
\subsection{Efficiency}
Table~\ref{tab:ef} compares the efficiency of MPRNet\cite{Mehri_Ardakani_Sappa_2021}, All-in-One \cite{Li_Tan_Cheong_2020}, BIDeN\cite{han2022bid}, CPNet\cite{Wang_2023_CVPR}, and our CBDNet. The metrics evaluated are the number of parameters and FLOPs. Among the methods compared, CBDNet has the fewest parameters and FLOPs.

\begin{table}[!htbp]
\centering
\fontsize{7}{3}\selectfont

\setlength{\tabcolsep}{2pt}
\caption{Comparison on the number of trainable parameters and FLOPs. The best performance is marked in \textbf{bold}. Our method significantly outperforms the comparative methods in both the amount of learnable parameters and FLOPs.}
\begin{tabular}{lccccc}
\toprule
Method &MPRNet &All-in-one &BIDeN &CPNet &Ours\cr

\midrule
Params (M)$\downarrow$ & 21.15 & 44.26 & 38.61 & 11.30 & \textbf{1.30}\cr
FLOPs (G)$\downarrow$ & 135 & 350 & 344 & 102 & \textbf{43}\cr
\bottomrule
\end{tabular}
\label{tab:ef}
\end{table}

\subsection{Task I: Multi-weather removal in driving}

\begin{figure}[!htb]
    \centering 
    \hspace*{-0.1\linewidth}
    \includegraphics[width=0.6\linewidth]{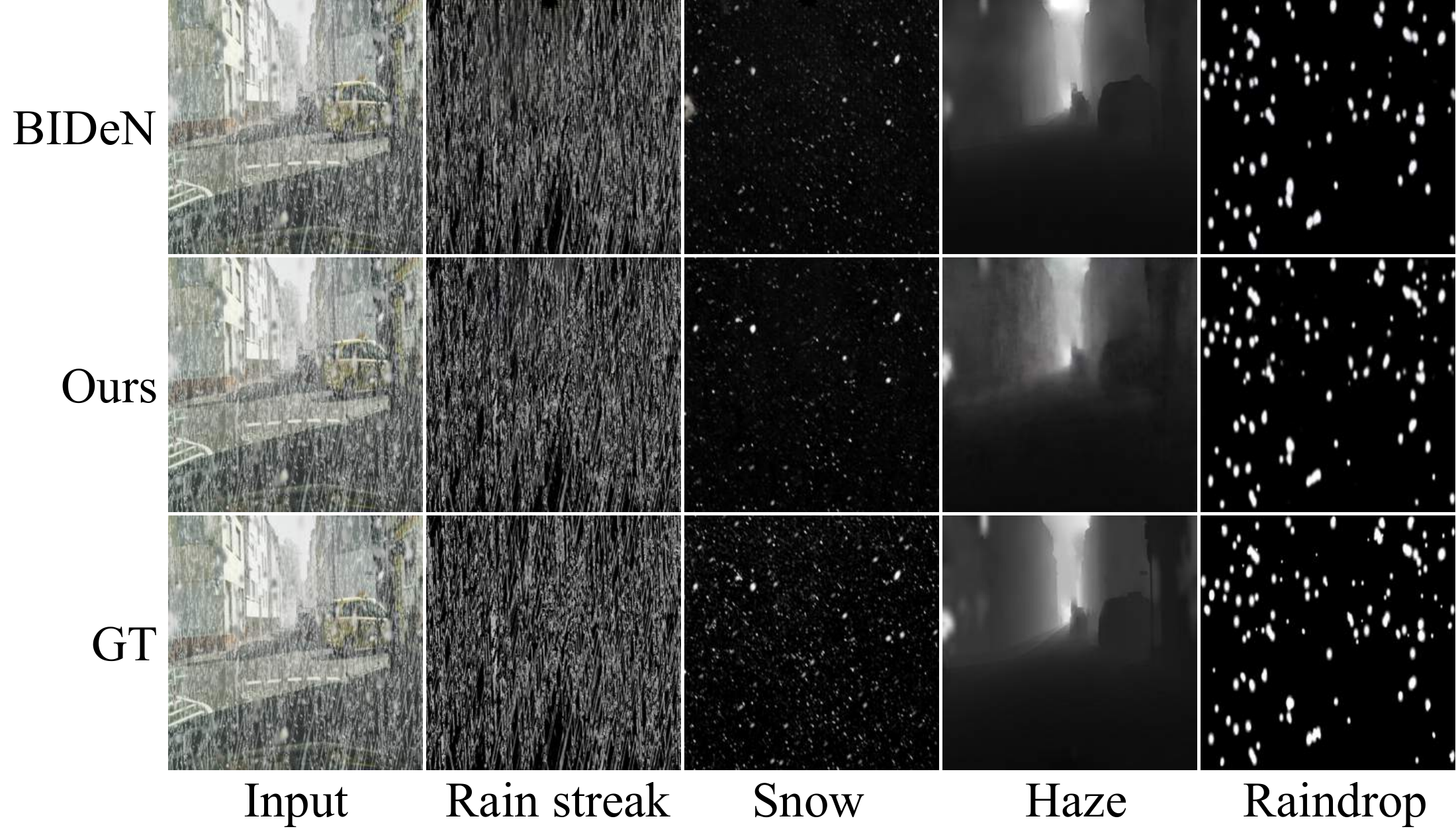}
\caption{Qualitative results of Task I.B (Masks reconstruction of multi-weather removal in driving). CBDNet exhibits comparable performance to BIDeN in haze reconstruction and demonstrates superior performance over BIDeN in the reconstruction of rain streak, snow, and raindrop.}
\label{fig:task1-mask-fig}
\end{figure}

\textbf{B: Masks reconstruction.} We compare the performance of BIDeN and CBDNet in mask reconstruction. We display the results of mask reconstruction for rain streak, snow, haze, and raindrop under the difficult case: rain streak + snow + moderate haze + raindrop in Table~\ref{tab:task1-b} and Figure~\ref{fig:task1-mask-fig}. In the reconstruction of haze, BIDeN demonstrates exceptional performance, with CBDNet exhibiting comparable performance. For the restoration of rain streaks, snow, and raindrops, CBDNet outperforms BIDeN across the board.

\begin{table}[!h]
\centering
\fontsize{7}{3}\selectfont
\setlength{\tabcolsep}{2pt}

\caption{Quantitative results on Task I.B (Degradation mask reconstruction of real-scenerio deraining in driving). We evaluate the performance of reconstruction of Rain-streak, Snow, Haze and Raindrop in PSNR and SSIM under the case: rain-streak + snow + moderate haze + raindrop. The best performance is marked in bold.}
\begin{tabular}{l|cc|cc|cc|cc}
\toprule
&\multicolumn{2}{c|}{Rain streak}&\multicolumn{2}{c|}{Snow}&\multicolumn{2}{c|}{Haze}&\multicolumn{2}{c}{Raindrop}\cr
Method& PSNR$\uparrow$ & SSIM$\uparrow$ & PSNR$\uparrow$ & SSIM$\uparrow$ & PSNR$\uparrow$ & SSIM$\uparrow$ & PSNR$\uparrow$ & SSIM$\uparrow$\cr
\midrule
BIDeN &28.32 &0.815 &24.80 &0.670 &29.83 &\textbf{0.942} &21.47 &0.893\\
\midrule
Ours &\textbf{29.68} &\textbf{0.818} &\textbf{27.02} &\textbf{0.800} &\textbf{30.93} &0.920 &\textbf{24.67} &\textbf{0.915}\cr
\bottomrule
\end{tabular}
\label{tab:task1-b}
\end{table}

\begin{figure}[!htb]
    \begin{minipage}{0.98\textwidth}
       	\begin{minipage}[c]{0.06\textwidth}
       	\hfill 
       	\text{\scriptsize Ours\;}
       \end{minipage}%
       \begin{minipage}[c]{0.155\textwidth}
       	\includegraphics[width=0.98\textwidth, height=0.49\textwidth]{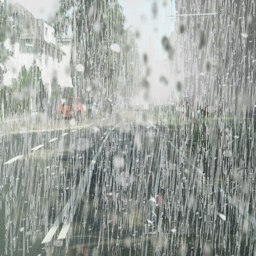}
       \end{minipage}%
       \begin{minipage}[c]{0.155\textwidth}
       	\includegraphics[width=0.98\textwidth, height=0.49\textwidth]{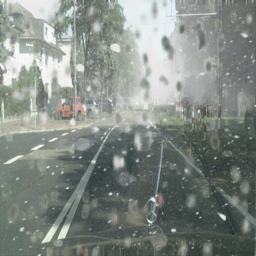}
       \end{minipage}%
       \begin{minipage}[c]{0.155\textwidth}
       	\includegraphics[width=0.98\textwidth, height=0.49\textwidth]{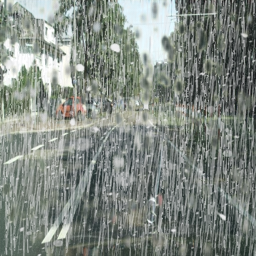}
       \end{minipage}%
       \begin{minipage}[c]{0.155\textwidth}
       	\includegraphics[width=0.98\textwidth, height=0.49\textwidth]{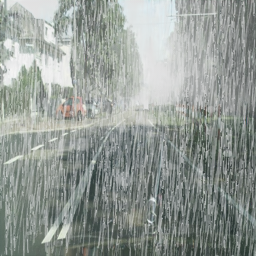}
       \end{minipage}%
       \begin{minipage}[c]{0.155\textwidth}
       	\includegraphics[width=0.98\textwidth, height=0.49\textwidth]{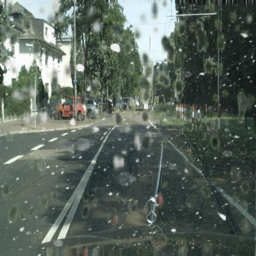}
       \end{minipage}%
        \begin{minipage}[c]{0.155\textwidth}
       	\includegraphics[width=0.98\textwidth, height=0.49\textwidth]{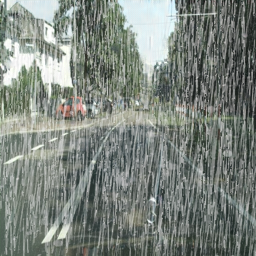}
       \end{minipage}%
   
        \begin{minipage}[c]{0.06\textwidth}
            \hfill 
            \text{\scriptsize GT\;}
        \end{minipage}%
        \begin{minipage}[c]{0.155\textwidth}
            \includegraphics[width=0.98\textwidth, height=0.49\textwidth]{figures/task1-c/1-0.png}
        \end{minipage}%
        \begin{minipage}[c]{0.155\textwidth}
            \includegraphics[width=0.98\textwidth, height=0.49\textwidth]{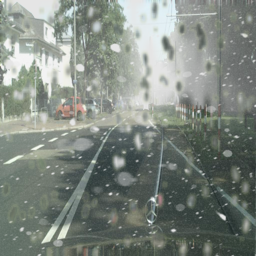}
        \end{minipage}%
        \begin{minipage}[c]{0.155\textwidth}
            \includegraphics[width=0.98\textwidth, height=0.49\textwidth]{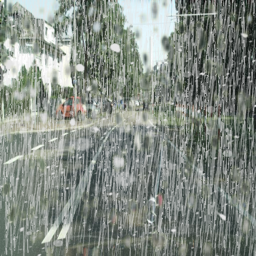}
        \end{minipage}%
        \begin{minipage}[c]{0.155\textwidth}
            \includegraphics[width=0.98\textwidth, height=0.49\textwidth]{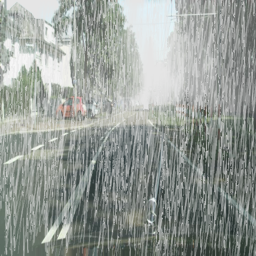}
        \end{minipage}%
        \begin{minipage}[c]{0.155\textwidth}
            \includegraphics[width=0.98\textwidth, height=0.49\textwidth]{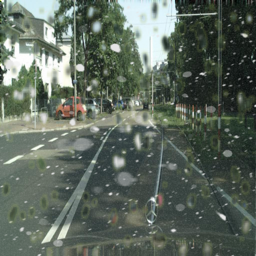}
        \end{minipage}%
        \begin{minipage}[c]{0.155\textwidth}
            \includegraphics[width=0.98\textwidth, height=0.49\textwidth]{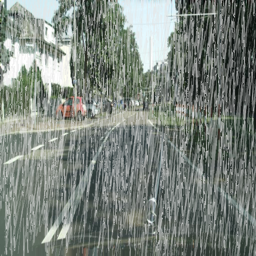}
        \end{minipage}%

        \begin{minipage}[c]{0.06\textwidth}
            \hfill 
            \text{\scriptsize Ours\;}
        \end{minipage}%
        \begin{minipage}[c]{0.155\textwidth}
            \includegraphics[width=0.98\textwidth, height=0.49\textwidth]{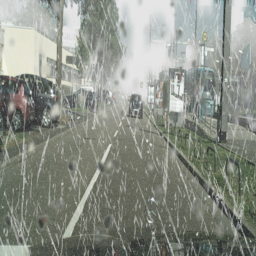}
        \end{minipage}%
        \begin{minipage}[c]{0.155\textwidth}
            \includegraphics[width=0.98\textwidth, height=0.49\textwidth]{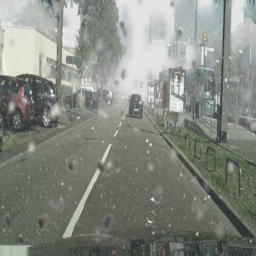}
        \end{minipage}%
        \begin{minipage}[c]{0.155\textwidth}
            \includegraphics[width=0.98\textwidth, height=0.49\textwidth]{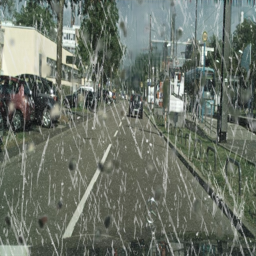}
        \end{minipage}%
        \begin{minipage}[c]{0.155\textwidth}
            \includegraphics[width=0.98\textwidth, height=0.49\textwidth]{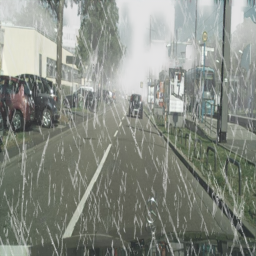}
        \end{minipage}%
        \begin{minipage}[c]{0.155\textwidth}
            \includegraphics[width=0.98\textwidth, height=0.49\textwidth]{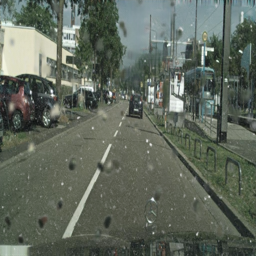}
        \end{minipage}%
        \begin{minipage}[c]{0.155\textwidth}
            \includegraphics[width=0.98\textwidth, height=0.49\textwidth]{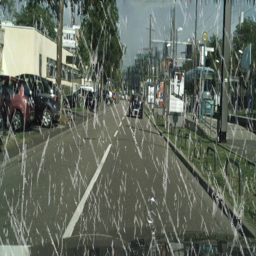}
        \end{minipage}%

        \begin{minipage}[c]{0.06\textwidth}
            \hfill 
            \text{\scriptsize GT\;}
        \end{minipage}%
        \begin{minipage}[c]{0.155\textwidth}
            \includegraphics[width=0.98\textwidth, height=0.49\textwidth]{figures/task1-c/2-0.png}
        \end{minipage}%
        \begin{minipage}[c]{0.155\textwidth}
            \includegraphics[width=0.98\textwidth, height=0.49\textwidth]{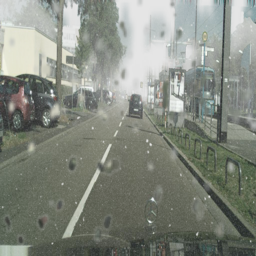}
        \end{minipage}%
        \begin{minipage}[c]{0.155\textwidth}
            \includegraphics[width=0.98\textwidth, height=0.49\textwidth]{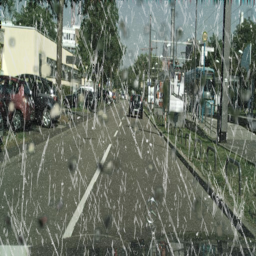}
        \end{minipage}%
        \begin{minipage}[c]{0.155\textwidth}
            \includegraphics[width=0.98\textwidth, height=0.49\textwidth]{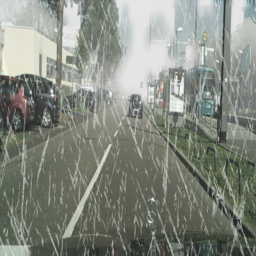}
        \end{minipage}%
        \begin{minipage}[c]{0.155\textwidth}
            \includegraphics[width=0.98\textwidth, height=0.49\textwidth]{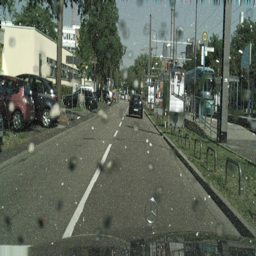}
        \end{minipage}%
        \begin{minipage}[c]{0.155\textwidth}
            \includegraphics[width=0.98\textwidth, height=0.49\textwidth]{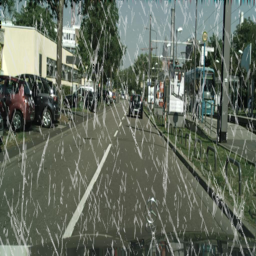}
        \end{minipage}%

        \begin{minipage}[c]{0.06\textwidth}
            \hfill 
            \text{\scriptsize Ours\;}
        \end{minipage}%
        \begin{minipage}[c]{0.155\textwidth}
            \includegraphics[width=0.98\textwidth, height=0.49\textwidth]{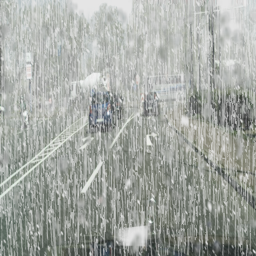}
        \end{minipage}%
        \begin{minipage}[c]{0.155\textwidth}
            \includegraphics[width=0.98\textwidth, height=0.49\textwidth]{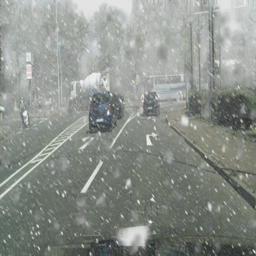}
        \end{minipage}%
        \begin{minipage}[c]{0.155\textwidth}
            \includegraphics[width=0.98\textwidth, height=0.49\textwidth]{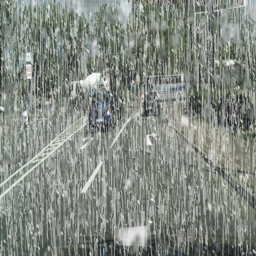}
        \end{minipage}%
        \begin{minipage}[c]{0.155\textwidth}
            \includegraphics[width=0.98\textwidth, height=0.49\textwidth]{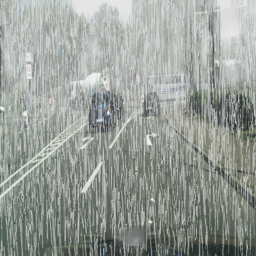}
        \end{minipage}%
        \begin{minipage}[c]{0.155\textwidth}
            \includegraphics[width=0.98\textwidth, height=0.49\textwidth]{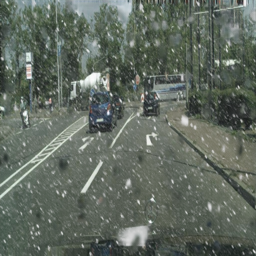}
        \end{minipage}%
        \begin{minipage}[c]{0.155\textwidth}
            \includegraphics[width=0.98\textwidth, height=0.49\textwidth]{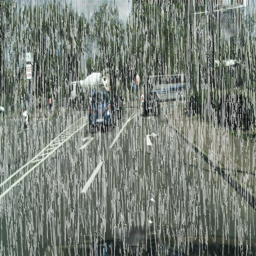}
        \end{minipage}%

        \begin{minipage}[c]{0.06\textwidth}
            \hfill 
            \text{\scriptsize GT\;}
        \end{minipage}%
        \begin{minipage}[c]{0.155\textwidth}
            \includegraphics[width=0.98\textwidth, height=0.49\textwidth]{figures/task1-c/3-0.png}
        \end{minipage}%
        \begin{minipage}[c]{0.155\textwidth}
            \includegraphics[width=0.98\textwidth, height=0.49\textwidth]{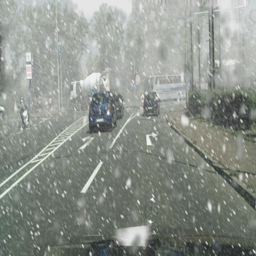}
        \end{minipage}%
        \begin{minipage}[c]{0.155\textwidth}
            \includegraphics[width=0.98\textwidth, height=0.49\textwidth]{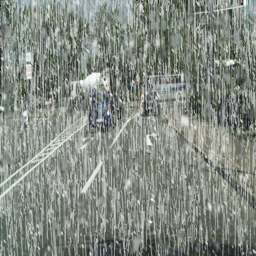}
        \end{minipage}%
        \begin{minipage}[c]{0.155\textwidth}
            \includegraphics[width=0.98\textwidth, height=0.49\textwidth]{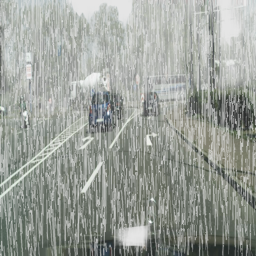}
        \end{minipage}%
        \begin{minipage}[c]{0.155\textwidth}
            \includegraphics[width=0.98\textwidth, height=0.49\textwidth]{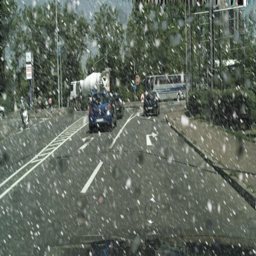}
        \end{minipage}%
        \begin{minipage}[c]{0.155\textwidth}
            \includegraphics[width=0.98\textwidth, height=0.49\textwidth]{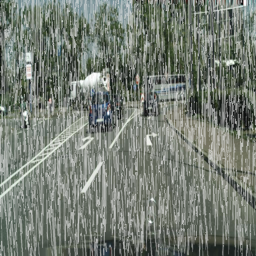}
        \end{minipage}%

        \begin{minipage}[c]{0.06\textwidth}
            \hfill 
            \text{\scriptsize Ours\;}
        \end{minipage}%
        \begin{minipage}[c]{0.155\textwidth}
            \includegraphics[width=0.98\textwidth, height=0.49\textwidth]{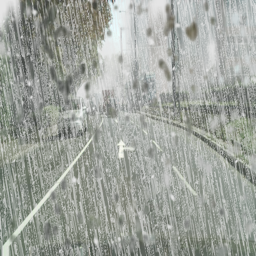}
        \end{minipage}%
        \begin{minipage}[c]{0.155\textwidth}
            \includegraphics[width=0.98\textwidth, height=0.49\textwidth]{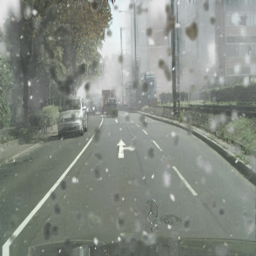}
        \end{minipage}%
        \begin{minipage}[c]{0.155\textwidth}
            \includegraphics[width=0.98\textwidth, height=0.49\textwidth]{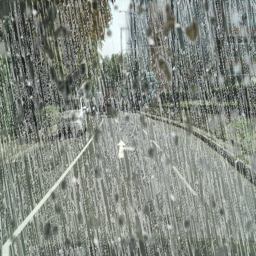}
        \end{minipage}%
        \begin{minipage}[c]{0.155\textwidth}
            \includegraphics[width=0.98\textwidth, height=0.49\textwidth]{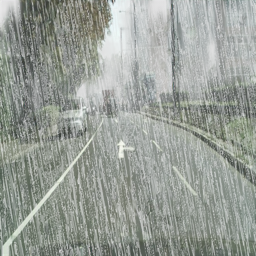}
        \end{minipage}%
        \begin{minipage}[c]{0.155\textwidth}
            \includegraphics[width=0.98\textwidth, height=0.49\textwidth]{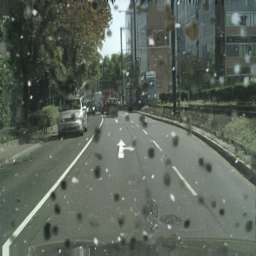}
        \end{minipage}%
        \begin{minipage}[c]{0.155\textwidth}
            \includegraphics[width=0.98\textwidth, height=0.49\textwidth]{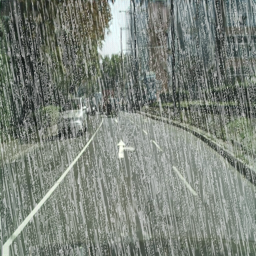}
        \end{minipage}%

        \begin{minipage}[c]{0.06\textwidth}
            \hfill 
            \text{\scriptsize GT\;}
        \end{minipage}%
        \begin{minipage}[c]{0.155\textwidth}
            \includegraphics[width=0.98\textwidth, height=0.49\textwidth]{figures/task1-c/4-0.png}
        \end{minipage}%
        \begin{minipage}[c]{0.155\textwidth}
            \includegraphics[width=0.98\textwidth, height=0.49\textwidth]{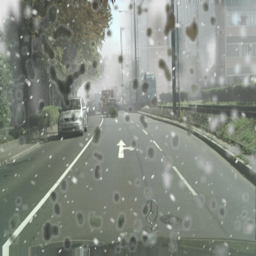}
        \end{minipage}%
        \begin{minipage}[c]{0.155\textwidth}
            \includegraphics[width=0.98\textwidth, height=0.49\textwidth]{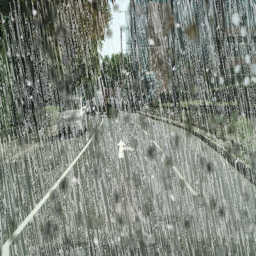}
        \end{minipage}%
        \begin{minipage}[c]{0.155\textwidth}
            \includegraphics[width=0.98\textwidth, height=0.49\textwidth]{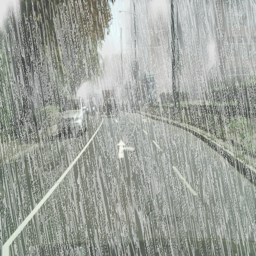}
        \end{minipage}%
        \begin{minipage}[c]{0.155\textwidth}
            \includegraphics[width=0.98\textwidth, height=0.49\textwidth]{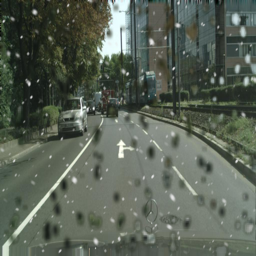}
        \end{minipage}%
        \begin{minipage}[c]{0.155\textwidth}
            \includegraphics[width=0.98\textwidth, height=0.49\textwidth]{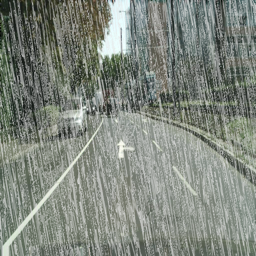}
        \end{minipage}%

        \begin{minipage}[c]{0.06\textwidth}
            \hfill 
            \text{\scriptsize \;}
        \end{minipage}%
        \begin{minipage}[c]{0.155\textwidth}
            \centering 
            \text{\scriptsize Input}
        \end{minipage}%
        \begin{minipage}[c]{0.155\textwidth}
            \centering 
            \text{\scriptsize Prompt (1)}
        \end{minipage}%
        \begin{minipage}[c]{0.155\textwidth}
            \centering 
            \text{\scriptsize Prompt (2)}
        \end{minipage}%
        \begin{minipage}[c]{0.155\textwidth}
            \centering 
            \text{\scriptsize Prompt (3)}
        \end{minipage}%
        \begin{minipage}[c]{0.155\textwidth}
            \centering 
            \text{\scriptsize Prompt (4)}
        \end{minipage}%
        \begin{minipage}[c]{0.155\textwidth}
            \centering 
            \text{\scriptsize Prompt (5)}
        \end{minipage}%

    \end{minipage}

    \caption{Qualitative results of Task I.C (Removing specific real-scenario degradation based on given prompt in driving). Prompt (1)-(5) are illustrated in Table~\ref{tab:task1-c}. Our method adeptly removes various degradations in response to specific prompts, resulting in generated images that are fundamentally consistent with the ground truth.}
    \label{fig:task1-c-fig}

\end{figure}

\noindent\textbf{C: Controllable Blind Image Decomposition.} We demonstrated the capability of our method in controllable blind image decomposition, which is to remove specific components based on text prompts. The results are shown in Table~\ref{tab:task1-c} and Figure~\ref{fig:task1-c-fig}. The input images consists of rain streak + snow + moderate haze + raindrop.

\begin{table}[!htbp]
\centering
\fontsize{7}{3}\selectfont
\setlength{\tabcolsep}{2pt}
\caption{Quantitative results on Task I.C (Controllable Blind Image Decomposition of real-scenario deraining in driving). All the inputs contain rain streak, snow, haze, and raindrop. Prompt (1): remove rain streak; Prompt (2): remove haze; Prompt (3): remove snow and raindrop; Prompt (4): remove rain streak and haze; Prompt (5): reconstruct the scene and rain streak.}
\begin{tabular}{l|c|c|c|c|c}
\toprule
\multicolumn{1}{c|}{Metric} & Prompt (1) & Prompt (2) & Prompt (3) & Prompt (4) & Prompt (5)\\
\midrule
PSNR$\uparrow$ & 36.10 & 30.70 & 32.41 & 29.02 & 29.29\\
SSIM$\uparrow$ & 0.962 & 0.969 & 0.938 & 0.928 & 0.919\\

\bottomrule
\end{tabular}

\label{tab:task1-c}
\end{table}

\subsection{Task II: Real-world bad weather removal}
\begin{figure}[!htb]
\centering 
\includegraphics[width=1\linewidth]{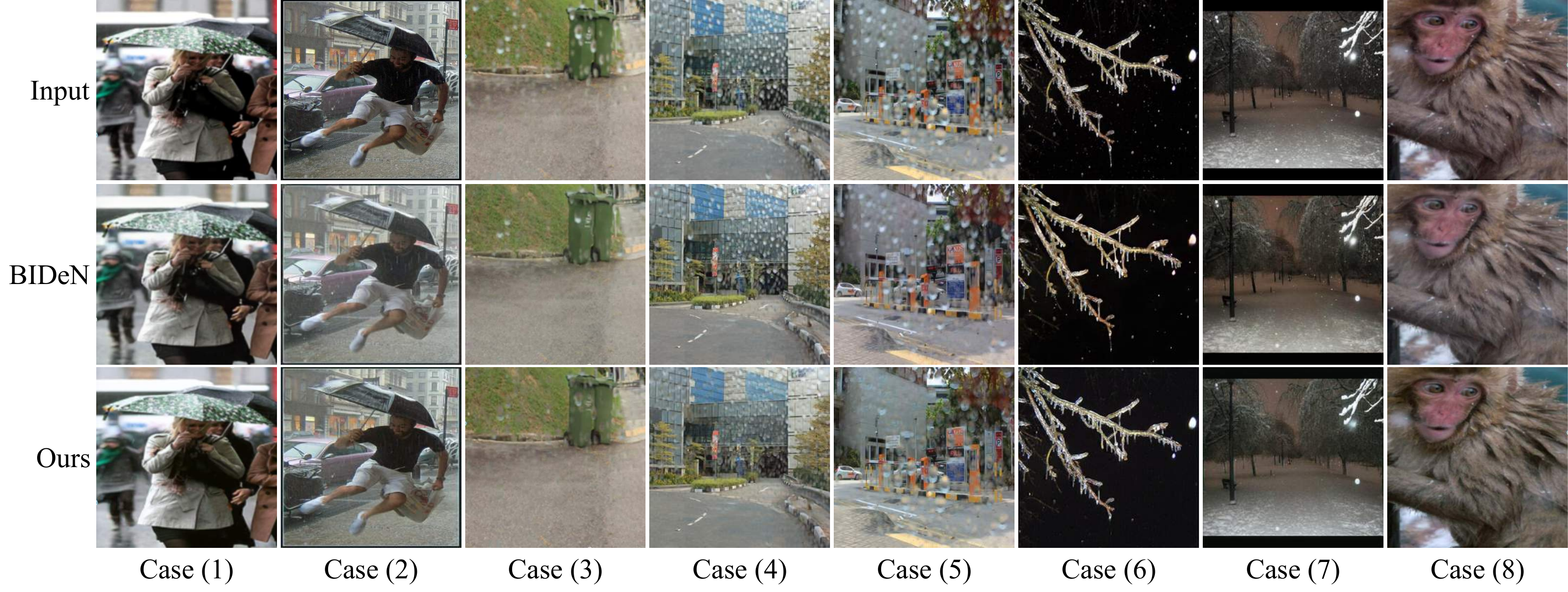}

\caption{Qualitative results of Task II: Real-world bad
weather removal. We showcase the performance of methods
across three real-world nature datasets, each encompassing distinct weather conditions: rain streak, raindrop, and snow. Cases (1) to (2) feature rain streaks; cases (3) to (5) showcase raindrops; cases (6) to (8) are characterized by snow. Please zoom in to view the details.}
  \label{fig:task2-supp-fig}
\end{figure}

In Figure~\ref{fig:task2-supp-fig}, we present the qualitative results of real-world bad weather removal.

\clearpage

\subsection{Task III: Multi-degradation removal}

\begin{figure}[!htb]
    \begin{minipage}{1\linewidth}
    \centering
        \begin{minipage}[c]{0.08\linewidth}
            \hfill 
            \text{\scriptsize Input\;}
        \end{minipage}%
        \begin{minipage}[c]{0.15\linewidth}
            \includegraphics[width=0.98\linewidth, height=0.49\linewidth]{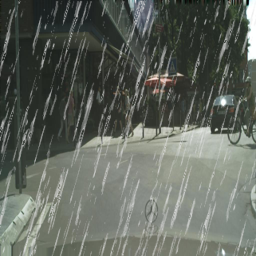}
        \end{minipage}%
        \begin{minipage}[c]{0.15\linewidth}
            \includegraphics[width=0.98\linewidth, height=0.49\linewidth]{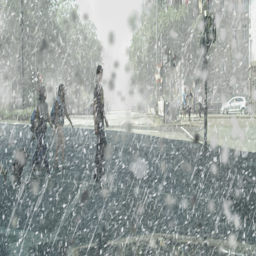}
        \end{minipage}%
        \begin{minipage}[c]{0.15\linewidth}
            \includegraphics[width=0.98\linewidth, height=0.49\linewidth]{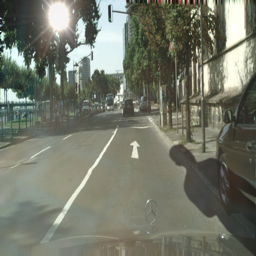}
        \end{minipage}%
        \begin{minipage}[c]{0.15\linewidth}
            \includegraphics[width=0.98\linewidth, height=0.49\linewidth]{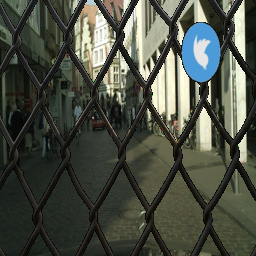}
        \end{minipage}%
        \begin{minipage}[c]{0.15\linewidth}
            \includegraphics[width=0.98\linewidth, height=0.49\linewidth]{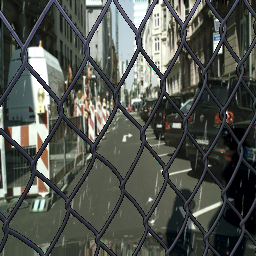}
        \end{minipage}%
        \begin{minipage}[c]{0.15\linewidth}
            \includegraphics[width=0.98\linewidth, height=0.49\linewidth]{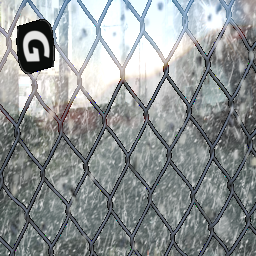}
        \end{minipage}%

        \begin{minipage}[c]{0.08\linewidth}
            \hfill 
            \text{\scriptsize BIDeN\;}
        \end{minipage}%
        \begin{minipage}[c]{0.15\linewidth}
            \includegraphics[width=0.98\linewidth, height=0.49\linewidth]{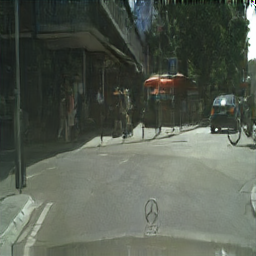}
        \end{minipage}%
        \begin{minipage}[c]{0.15\linewidth}
            \includegraphics[width=0.98\linewidth, height=0.49\linewidth]{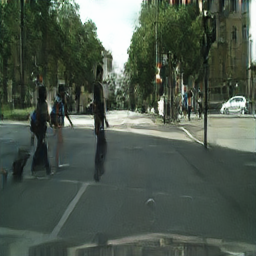}
        \end{minipage}%
        \begin{minipage}[c]{0.15\linewidth}
            \includegraphics[width=0.98\linewidth, height=0.49\linewidth]{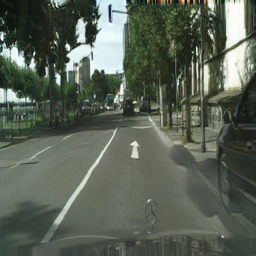}
        \end{minipage}%
        \begin{minipage}[c]{0.15\linewidth}
            \includegraphics[width=0.98\linewidth, height=0.49\linewidth]{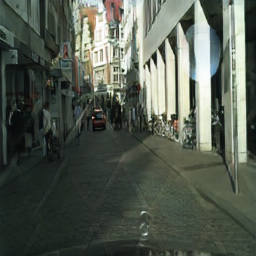}
        \end{minipage}%
        \begin{minipage}[c]{0.15\linewidth}
            \includegraphics[width=0.98\linewidth, height=0.49\linewidth]{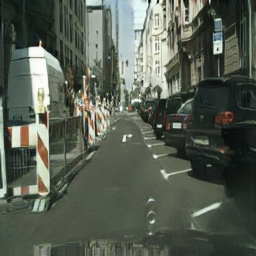}
        \end{minipage}%
        \begin{minipage}[c]{0.15\linewidth}
            \includegraphics[width=0.98\linewidth, height=0.49\linewidth]{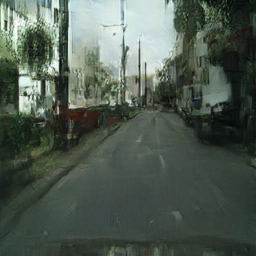}
        \end{minipage}%

        \begin{minipage}[c]{0.08\linewidth}
            \hfill 
            \text{\scriptsize Ours\;}
        \end{minipage}%
        \begin{minipage}[c]{0.15\linewidth}
            \includegraphics[width=0.98\linewidth, height=0.49\linewidth]{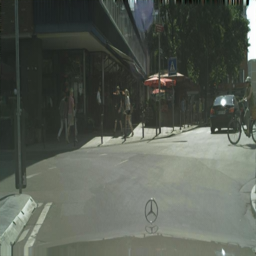}
        \end{minipage}%
        \begin{minipage}[c]{0.15\linewidth}
            \includegraphics[width=0.98\linewidth, height=0.49\linewidth]{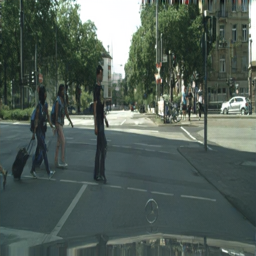}
        \end{minipage}%
        \begin{minipage}[c]{0.15\linewidth}
            \includegraphics[width=0.98\linewidth, height=0.49\linewidth]{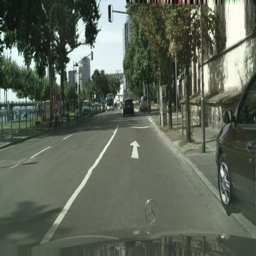}
        \end{minipage}%
        \begin{minipage}[c]{0.15\linewidth}
            \includegraphics[width=0.98\linewidth, height=0.49\linewidth]{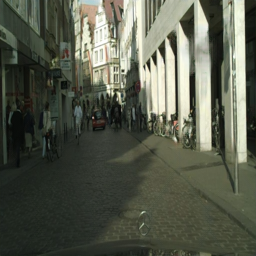}
        \end{minipage}%
        \begin{minipage}[c]{0.15\linewidth}
            \includegraphics[width=0.98\linewidth, height=0.49\linewidth]{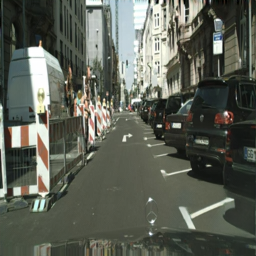}
        \end{minipage}%
        \begin{minipage}[c]{0.15\linewidth}
            \includegraphics[width=0.98\linewidth, height=0.49\linewidth]{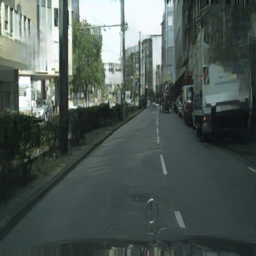}
        \end{minipage}%

        \begin{minipage}[c]{0.08\linewidth}
            \hfill 
            \text{\scriptsize GT\;}
        \end{minipage}%
        \begin{minipage}[c]{0.15\linewidth}
            \includegraphics[width=0.98\linewidth, height=0.49\linewidth]{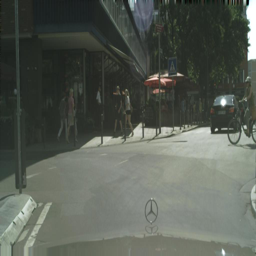}
        \end{minipage}%
        \begin{minipage}[c]{0.15\linewidth}
            \includegraphics[width=0.98\linewidth, height=0.49\linewidth]{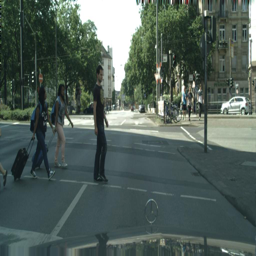}
        \end{minipage}%
        \begin{minipage}[c]{0.15\linewidth}
            \includegraphics[width=0.98\linewidth, height=0.49\linewidth]{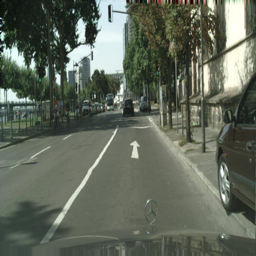}
        \end{minipage}%
        \begin{minipage}[c]{0.15\linewidth}
            \includegraphics[width=0.98\linewidth, height=0.49\linewidth]{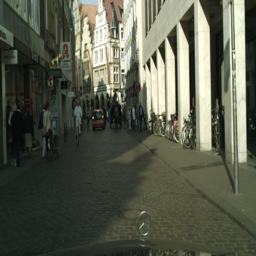}
        \end{minipage}%
        \begin{minipage}[c]{0.15\linewidth}
            \includegraphics[width=0.98\linewidth, height=0.49\linewidth]{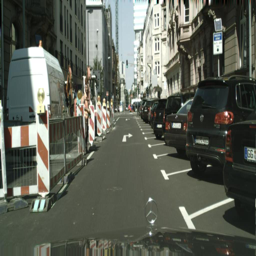}
        \end{minipage}%
        \begin{minipage}[c]{0.15\linewidth}
            \includegraphics[width=0.98\linewidth, height=0.49\linewidth]{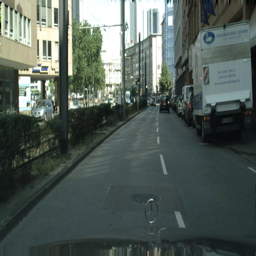}
        \end{minipage}%

        \begin{minipage}[c]{0.08\linewidth}
            \hfill 
            \text{\scriptsize \;}
        \end{minipage}%
        \begin{minipage}[c]{0.15\linewidth}
            \centering 
            \text{\scriptsize Case (1)}
        \end{minipage}%
        \begin{minipage}[c]{0.15\linewidth}
            \centering 
            \text{\scriptsize Case (2)}
        \end{minipage}%
        \begin{minipage}[c]{0.15\linewidth}
            \centering 
            \text{\scriptsize Case (3)}
        \end{minipage}%
        \begin{minipage}[c]{0.15\linewidth}
            \centering 
            \text{\scriptsize Case (4)}
        \end{minipage}%
        \begin{minipage}[c]{0.15\linewidth}
            \centering 
            \text{\scriptsize Case (5)}
        \end{minipage}%
        \begin{minipage}[c]{0.15\linewidth}
            \centering 
            \text{\scriptsize Case (6)}
        \end{minipage}%

    \end{minipage}

    \caption{Additional qualitative results of Task III.A : Image restoration in multi-degradation removal. Cases are: (1) rain streak; (2) rain streak + snow + moderate haze + raindrop; (3) flare + reflection + shadow; (4) fence + watermark; (5) rain streak + shadow + fence; (6) rain streak + snow + moderate haze + raindrop + flare + shadow + fence + reflection + watermark. Ours proposed CBDNet surpasses BIDeN in all cases, effectively restoring images with low visibility.}
    \label{fig:task3-a-supp-fig}

\end{figure}

\textbf{A: Image restoration.}
We report additional results for 6 cases in Figure~\ref{fig:task3-a-supp-fig}. 

\begin{figure}[!htb]
    \centering %
    \includegraphics[width=1\linewidth]{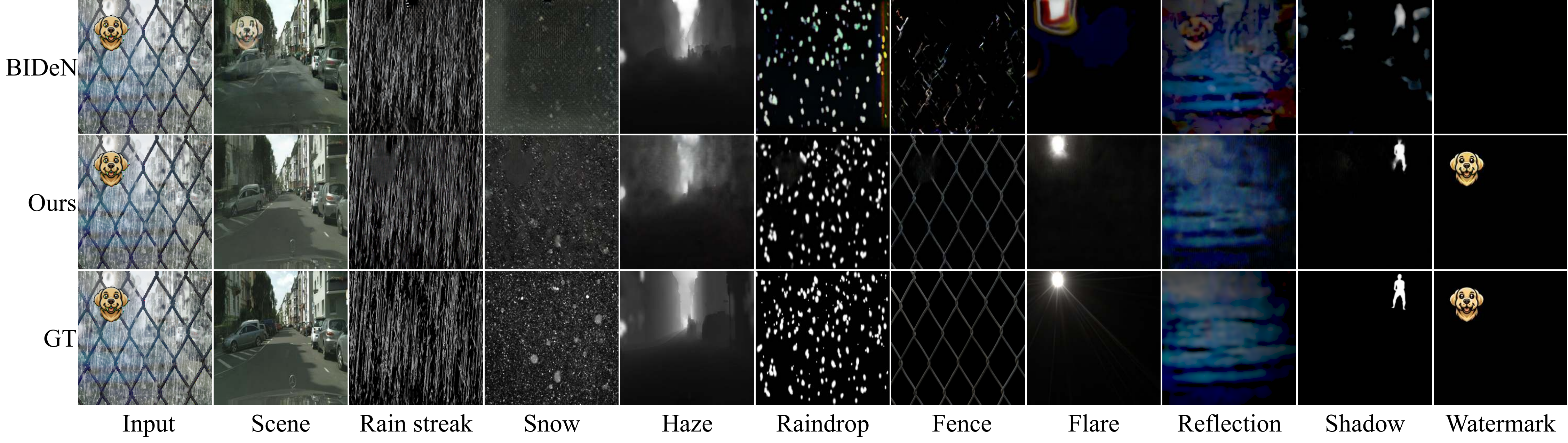}
    \caption{Additional qualitative results of Task III.B: Degradation masks reconstruction in multi-degradation removel. The input images contains all nine degradations. BIDeN is not able to effectively separate the various degradation components, which is reflected in the failed reconstruction of the degradation masks, such as snow, flare and shadow. Our proposed CBDNet successfully separates and reconstructs all components.}
    \label{fig:task3-b-supp-fig}
\end{figure}

\noindent\textbf{B: Degradation Masks Reconstruction.} We show additional results of mask reconstruction in the case where all nine degradations are presented in Figure~\ref{fig:task3-b-supp-fig}. 

\begin{figure*}[!b]
    \begin{minipage}{1\textwidth}
    \centering
        \begin{minipage}[c]{0.08\linewidth}
            \hfill 
            \text{\scriptsize Input\;}
        \end{minipage}%
        \begin{minipage}[c]{0.13\linewidth}
            \includegraphics[width=0.96\linewidth, height=0.48\linewidth]{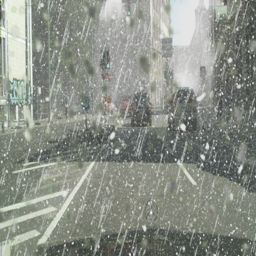}
        \end{minipage}%
        \begin{minipage}[c]{0.13\linewidth}
            \includegraphics[width=0.96\linewidth, height=0.48\linewidth]{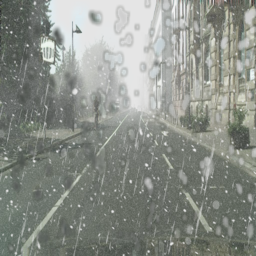}
        \end{minipage}%
        \begin{minipage}[c]{0.13\linewidth}
            \includegraphics[width=0.96\linewidth, height=0.48\linewidth]{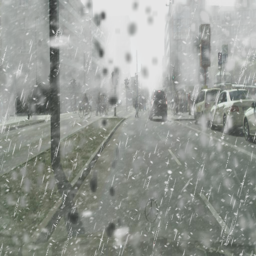}
        \end{minipage}%
        \begin{minipage}[c]{0.13\linewidth}
            \includegraphics[width=0.96\linewidth, height=0.48\linewidth]{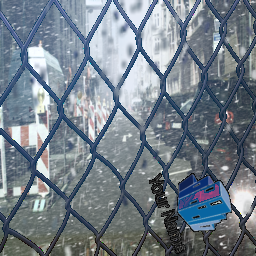}
        \end{minipage}%
        \begin{minipage}[c]{0.13\linewidth}
            \includegraphics[width=0.96\linewidth, height=0.48\linewidth]{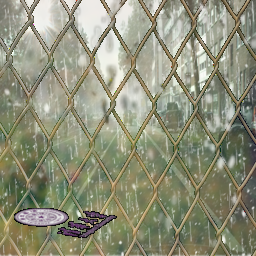}
        \end{minipage}%
        \begin{minipage}[c]{0.13\linewidth}
            \includegraphics[width=0.96\linewidth, height=0.48\linewidth]{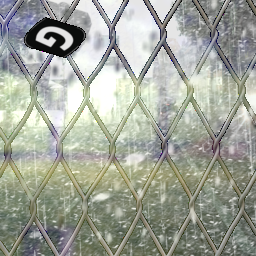}
        \end{minipage}%
        \begin{minipage}[c]{0.13\linewidth}
            \includegraphics[width=0.96\linewidth, height=0.48\linewidth]{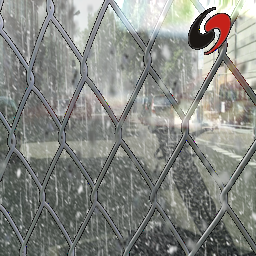}
        \end{minipage}%
        
        \begin{minipage}[c]{0.08\linewidth}
            \hfill 
            \text{\scriptsize Ours\;}
        \end{minipage}%
        \begin{minipage}[c]{0.13\linewidth}
            \includegraphics[width=0.96\linewidth, height=0.48\linewidth]{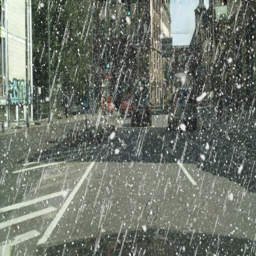}
        \end{minipage}%
        \begin{minipage}[c]{0.13\linewidth}
            \includegraphics[width=0.96\linewidth, height=0.48\linewidth]{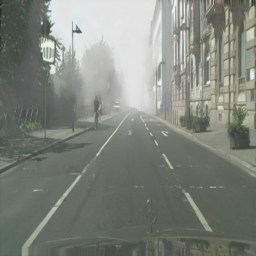}
        \end{minipage}%
        \begin{minipage}[c]{0.13\linewidth}
            \includegraphics[width=0.96\linewidth, height=0.48\linewidth]{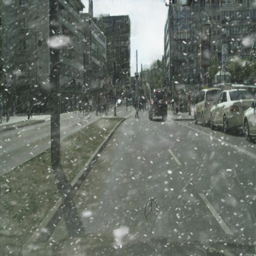}
        \end{minipage}%
        \begin{minipage}[c]{0.13\linewidth}
            \includegraphics[width=0.96\linewidth, height=0.48\linewidth]{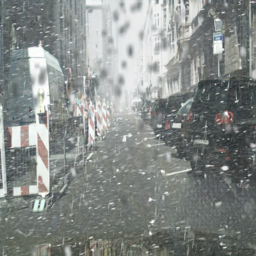}
        \end{minipage}%
        \begin{minipage}[c]{0.13\linewidth}
            \includegraphics[width=0.96\linewidth, height=0.48\linewidth]{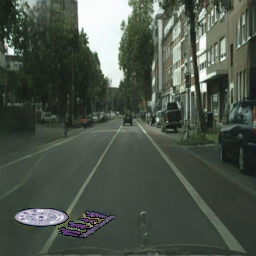}
        \end{minipage}%
        \begin{minipage}[c]{0.13\linewidth}
            \includegraphics[width=0.96\linewidth, height=0.48\linewidth]{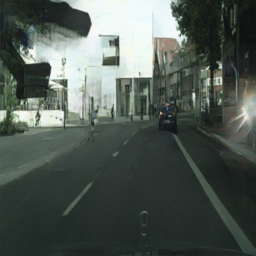}
        \end{minipage}%
        \begin{minipage}[c]{0.13\linewidth}
            \includegraphics[width=0.96\linewidth, height=0.48\linewidth]{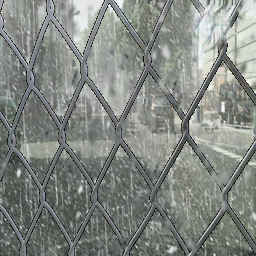}
        \end{minipage}%

        \begin{minipage}[c]{0.08\linewidth}
            \hfill 
            \text{\scriptsize GT\;}
        \end{minipage}%
        \begin{minipage}[c]{0.13\linewidth}
            \includegraphics[width=0.96\linewidth, height=0.48\linewidth]{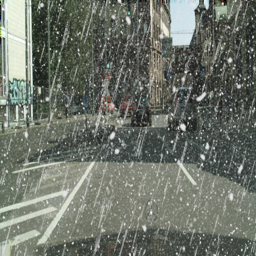}
        \end{minipage}%
        \begin{minipage}[c]{0.13\linewidth}
            \includegraphics[width=0.96\linewidth, height=0.48\linewidth]{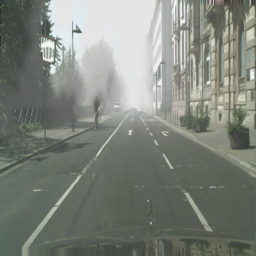}
        \end{minipage}%
        \begin{minipage}[c]{0.13\linewidth}
            \includegraphics[width=0.96\linewidth, height=0.48\linewidth]{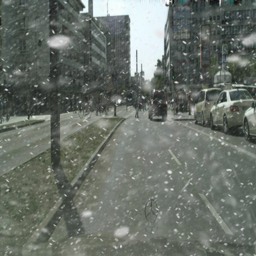}
        \end{minipage}%
        \begin{minipage}[c]{0.13\linewidth}
            \includegraphics[width=0.96\linewidth, height=0.48\linewidth]{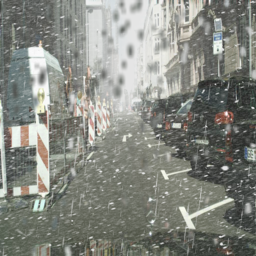}
        \end{minipage}%
        \begin{minipage}[c]{0.13\linewidth}
            \includegraphics[width=0.96\linewidth, height=0.48\linewidth]{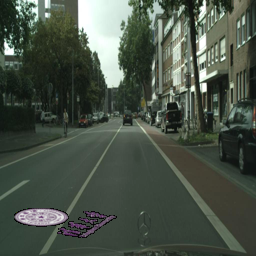}
        \end{minipage}%
        \begin{minipage}[c]{0.13\linewidth}
            \includegraphics[width=0.96\linewidth, height=0.48\linewidth]{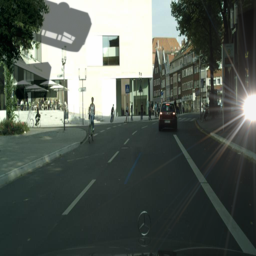}
        \end{minipage}%
        \begin{minipage}[c]{0.13\linewidth}
            \includegraphics[width=0.96\linewidth, height=0.48\linewidth]{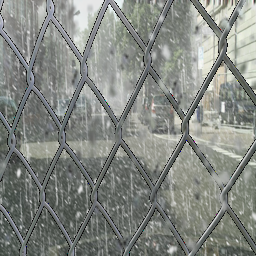}
        \end{minipage}%

        \begin{minipage}[c]{0.08\linewidth}
            \hfill 
            \text{\scriptsize Input\;}
        \end{minipage}%
        \begin{minipage}[c]{0.13\linewidth}
            \includegraphics[width=0.96\linewidth, height=0.48\linewidth]{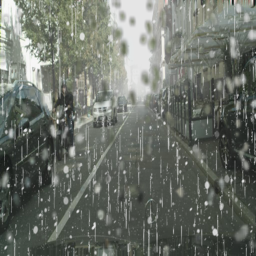}
        \end{minipage}%
        \begin{minipage}[c]{0.13\linewidth}
            \includegraphics[width=0.96\linewidth, height=0.48\linewidth]{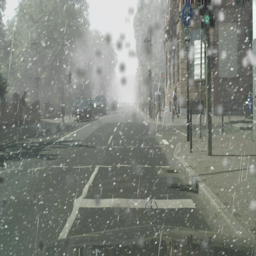}
        \end{minipage}%
        \begin{minipage}[c]{0.13\linewidth}
            \includegraphics[width=0.96\linewidth, height=0.48\linewidth]{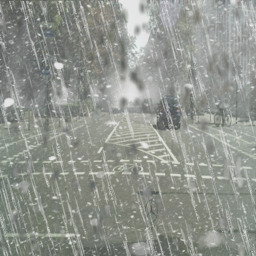}
        \end{minipage}%
        \begin{minipage}[c]{0.13\linewidth}
            \includegraphics[width=0.96\linewidth, height=0.48\linewidth]{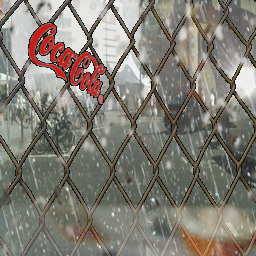}
        \end{minipage}%
        \begin{minipage}[c]{0.13\linewidth}
            \includegraphics[width=0.96\linewidth, height=0.48\linewidth]{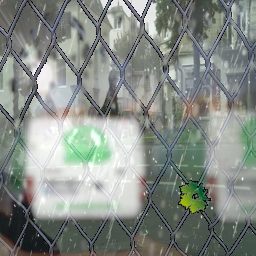}
        \end{minipage}%
        \begin{minipage}[c]{0.13\linewidth}
            \includegraphics[width=0.96\linewidth, height=0.48\linewidth]{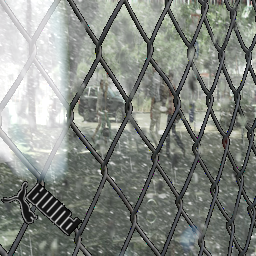}
        \end{minipage}%
        \begin{minipage}[c]{0.13\linewidth}
            \includegraphics[width=0.96\linewidth, height=0.48\linewidth]{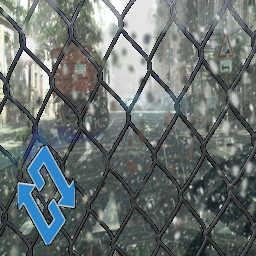}
        \end{minipage}%
        
        \begin{minipage}[c]{0.08\linewidth}
            \hfill 
            \text{\scriptsize Ours\;}
        \end{minipage}%
        \begin{minipage}[c]{0.13\linewidth}
            \includegraphics[width=0.96\linewidth, height=0.48\linewidth]{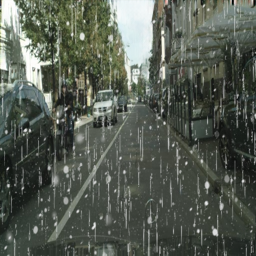}
        \end{minipage}%
        \begin{minipage}[c]{0.13\linewidth}
            \includegraphics[width=0.96\linewidth, height=0.48\linewidth]{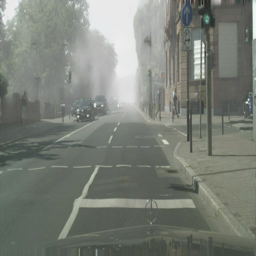}
        \end{minipage}%
        \begin{minipage}[c]{0.13\linewidth}
            \includegraphics[width=0.96\linewidth, height=0.48\linewidth]{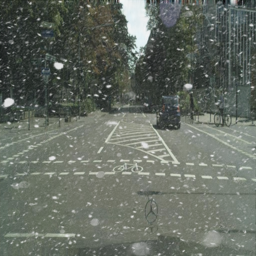}
        \end{minipage}%
        \begin{minipage}[c]{0.13\linewidth}
            \includegraphics[width=0.96\linewidth, height=0.48\linewidth]{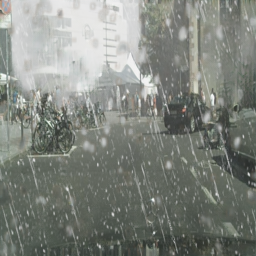}
        \end{minipage}%
        \begin{minipage}[c]{0.13\linewidth}
            \includegraphics[width=0.96\linewidth, height=0.48\linewidth]{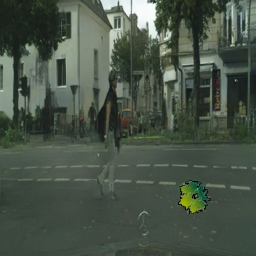}
        \end{minipage}%
        \begin{minipage}[c]{0.13\linewidth}
            \includegraphics[width=0.96\linewidth, height=0.48\linewidth]{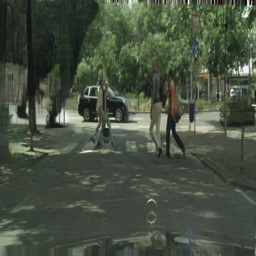}
        \end{minipage}%
        \begin{minipage}[c]{0.13\linewidth}
            \includegraphics[width=0.96\linewidth, height=0.48\linewidth]{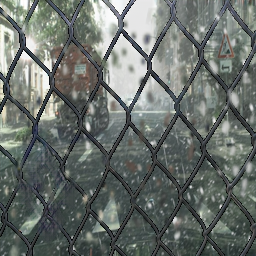}
        \end{minipage}%

        \begin{minipage}[c]{0.08\linewidth}
            \hfill 
            \text{\scriptsize GT\;}
        \end{minipage}%
        \begin{minipage}[c]{0.13\linewidth}
            \includegraphics[width=0.96\linewidth, height=0.48\linewidth]{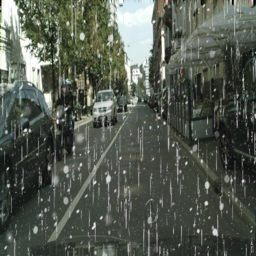}
        \end{minipage}%
        \begin{minipage}[c]{0.13\linewidth}
            \includegraphics[width=0.96\linewidth, height=0.48\linewidth]{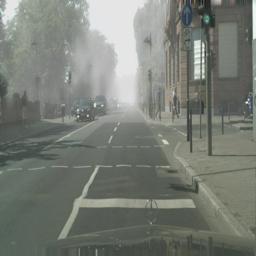}
        \end{minipage}%
        \begin{minipage}[c]{0.13\linewidth}
            \includegraphics[width=0.96\linewidth, height=0.48\linewidth]{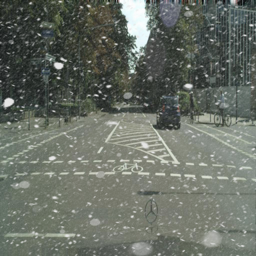}
        \end{minipage}%
        \begin{minipage}[c]{0.13\linewidth}
            \includegraphics[width=0.96\linewidth, height=0.48\linewidth]{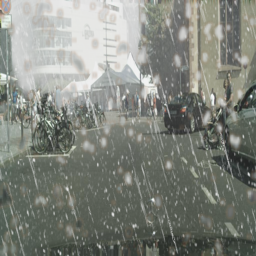}
        \end{minipage}%
        \begin{minipage}[c]{0.13\linewidth}
            \includegraphics[width=0.96\linewidth, height=0.48\linewidth]{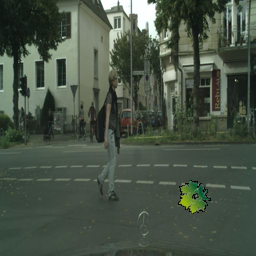}
        \end{minipage}%
        \begin{minipage}[c]{0.13\linewidth}
            \includegraphics[width=0.96\linewidth, height=0.48\linewidth]{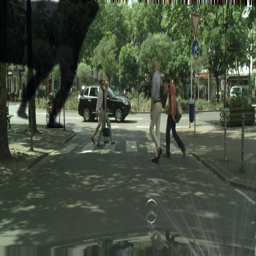}
        \end{minipage}%
        \begin{minipage}[c]{0.13\linewidth}
            \includegraphics[width=0.96\linewidth, height=0.48\linewidth]{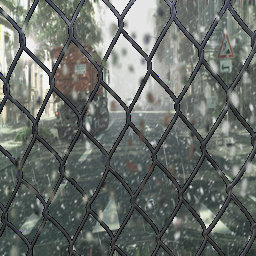}
        \end{minipage}%

        \begin{minipage}[c]{0.08\linewidth}
            \hfill 
            \text{\scriptsize \;}
        \end{minipage}%
        \begin{minipage}[c]{0.13\linewidth}
            \centering 
            \text{\scriptsize Prompt (1)}
        \end{minipage}%
        \begin{minipage}[c]{0.13\linewidth}
            \centering 
            \text{\scriptsize Prompt (2)}
        \end{minipage}%
        \begin{minipage}[c]{0.13\linewidth}
            \centering 
            \text{\scriptsize Prompt (3)}
        \end{minipage}%
        \begin{minipage}[c]{0.13\linewidth}
            \centering 
            \text{\scriptsize Prompt (4)}
        \end{minipage}%
        \begin{minipage}[c]{0.13\linewidth}
            \centering 
            \text{\scriptsize Prompt (5)}
        \end{minipage}%
        \begin{minipage}[c]{0.13\linewidth}
            \centering 
            \text{\scriptsize Prompt (6)}
        \end{minipage}%
        \begin{minipage}[c]{0.13\linewidth}
            \centering 
            \text{\scriptsize Prompt (7)}
        \end{minipage}%
    \end{minipage}

    \caption{Additional qualitative results of Task III.C: Controllable blind image decomposition in multi-degradation removal. Prompt (1): remove the haze and the raindrop; Prompt (2): generate the foggy scene; Prompt (3): retain the scene image and the snow; Prompt (4): keep the background and all weather components; Prompt (5): compose image with watermark; Prompt (6): restore scene, flare and shadow together; Prompt (7): remove watermark, reflection and shadow.}
    \label{fig:task3-c-supp-fig}
\end{figure*}

\noindent\textbf{C: Controllable blind image decomposition.}
We demonstrate the capability of our method to address specific types of degradation in Figure~\ref{fig:task3-c-supp-fig}.

\end{document}